%% file: main.tex
\documentclass[11pt,a4paper,copyright]{google}

\usepackage[authoryear,sort&compress,round]{natbib}
\usepackage{xcolor}
\definecolor{linkblue}{RGB}{0,0,139}
\definecolor{royalblue}{RGB}{65,105,225}
\usepackage[colorlinks=true,linkcolor=linkblue,urlcolor=royalblue,citecolor=brown,anchorcolor=blue]{hyperref}
\hypersetup{
  pdftitle={PAST-Bench: Benchmarking the Foundations of Recursive Self-Improvement in Personal Agents},
  pdfauthor={Shuhan Xue, Zixin Ding, Yichen Shen, Yinjie Wang, Zhenfei Yin, Yingcheng Wu, Yuxin Chen, Mengdi Wang, Ling Yang}
}
\usepackage{booktabs}
\usepackage{nicefrac}
\usepackage{multirow}
\usepackage{wrapfig}
\usepackage{graphicx}
\usepackage{caption}
\usepackage{xspace}
\usepackage{pifont}
\usepackage{amsmath}
\usepackage{tabularx}
\usepackage{longtable}
\usepackage{enumitem}
\usepackage[titles]{tocloft}
\usepackage[most]{tcolorbox}
\usepackage{float}

\makeatletter
\renewcommand{\absfont}{\normalfont\linespread{1.2}\fontsize{11}{12}\selectfont}
\makeatother

\newcommand{\benchname}{\textsc{PAST-Bench}\xspace}
\newcommand{\ouragent}{\textsc{Hermes+}\xspace}
\newcommand{\dpct}[1]{{\scriptscriptstyle\,(#1\%)}}

\newcommand{\benchyes}{\textcolor{green!65!black}{\ding{51}}}
\newcommand{\benchno}{\textcolor{red!75!black}{\ding{55}}}
\newcommand{\benchdash}{\textcolor{gray}{--}}
\newcommand{\benchpartial}{\textcolor{gray}{\(\triangle\)}}

\newcommand{\answerTODO}[1][]{\textcolor{red}{\bfseries [TODO]}}
\newcommand{\justificationTODO}[1][]{\textcolor{red}{\bfseries [TODO]}}

\uselogo{}
\title{PAST-Bench: Benchmarking the Foundations of Recursive Self-Improvement in Personal Agents}
\correspondingauthor{yangling0818@163.com}

\author{\quad Shuhan Xue$^*$, Zixin Ding$^*$, Yichen Shen$^*$, Yinjie Wang, Zhenfei Yin, Yingcheng Wu,\quad   Yuxin Chen,
Mengdi Wang$^\dagger$, Ling Yang$^\dagger$}

\makeatletter
\renewcommand{\maketitle}{\bgroup\setlength{\parindent}{0pt}
  \begin{adjustwidth}{0pt}{24pt}
    \begin{center}
      {\titlefont \@title\par}%
      \vskip11pt
      {\@author\par}%
    \end{center}
  \end{adjustwidth}
  \egroup
  \begin{center}
    {\fontsize{11pt}{13pt}\selectfont
     \raisebox{-0.06em}{\includegraphics[height=1em]{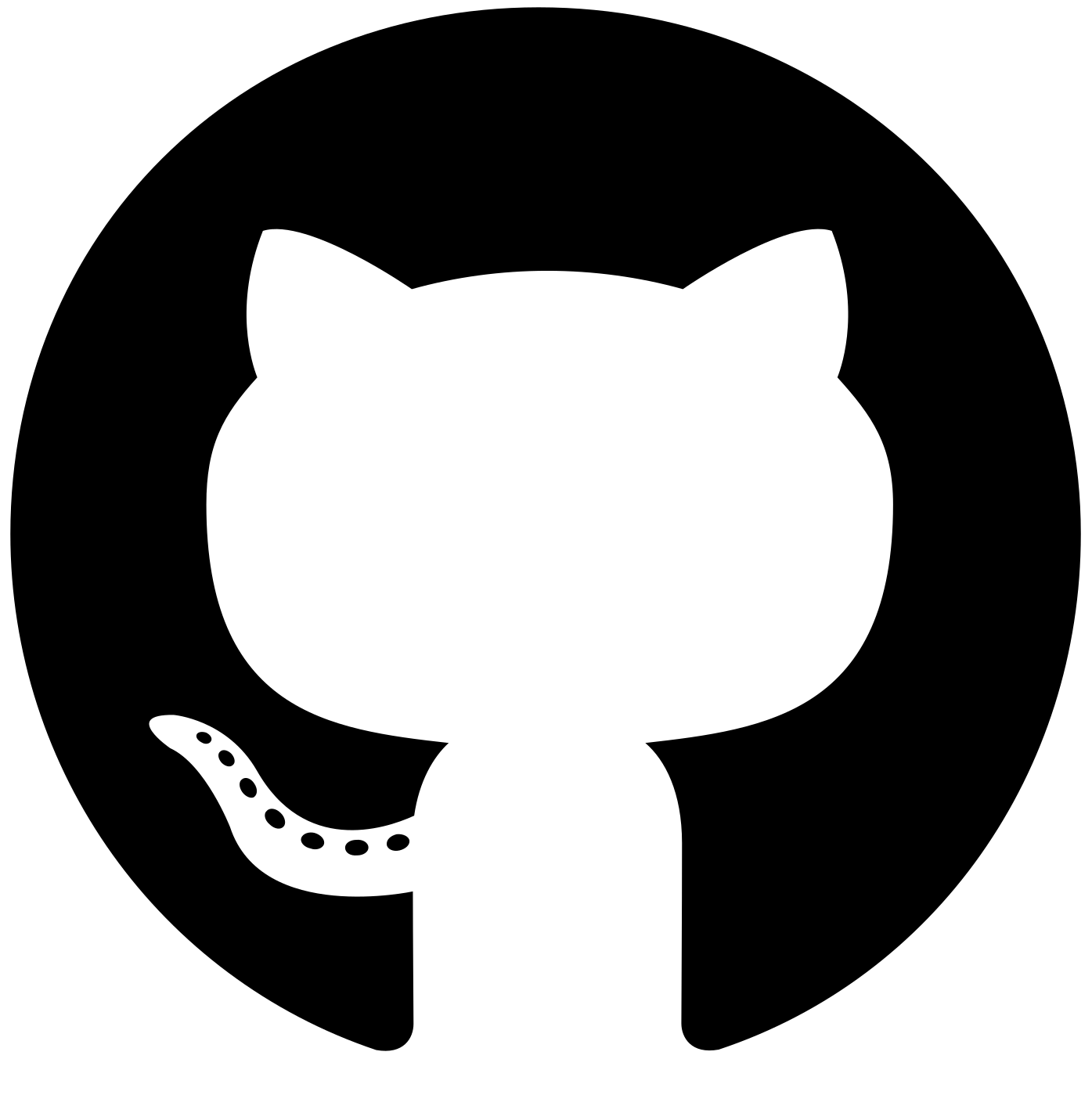}}\;
     \href{https://github.com/Gen-Verse/PAST-Bench}{{\fontfamily{lmtt}\fontsize{12.1pt}{13pt}\selectfont https://github.com/Gen-Verse/PAST-Bench}}}
  \end{center}
  {\abscontent}%
  \thispagestyle{firststyle}
}
\makeatother

\begin{abstract} 
Recursive self-improvement requires agents to turn accumulated experience into better future behavior. Personal AI agents offer a concrete setting for studying this capability because they retain preferences, task histories, tool routines, and learned skills across sessions. Yet whether retained experience actually improves them over time has not been systematically tested. We introduce \textbf{PAST-Bench}, a benchmark designed to isolate this question. Each agent runs through ordered sequences of fresh-session tasks under matched conditions that turn retained experience on and off. It spans \textbf{26 scenarios and 204 episodes} across memory, procedural reuse, information gathering, and update. We report both later-task gains and whether those gains follow the intended save, retrieve, and update pathway. Across seven base models and four agent frameworks, improvement is real but uneven across capabilities. Agents with the same headline gain can differ markedly in whether that gain is supported by evidence of the intended pathway. Guided by these findings, we develop \textbf{Hermes+}, which extends Hermes with five targeted interventions across stages of the agent loop. \textsc{Hermes+} raises the average gain from retained experience and provides clearer pathway evidence, with its strongest improvement on tasks requiring outdated state to be replaced, although the effect remains capability- and model-dependent. Together, \textbf{PAST-Bench} and \textbf{Hermes+} provide an evaluation and diagnostic foundation for studying how persistent agents can progress from retaining experience to systematically improving through it.
\end{abstract}

\begin{document}
\maketitle

\begin{figure}[H]
  \centering
  \includegraphics[width=0.99\linewidth]{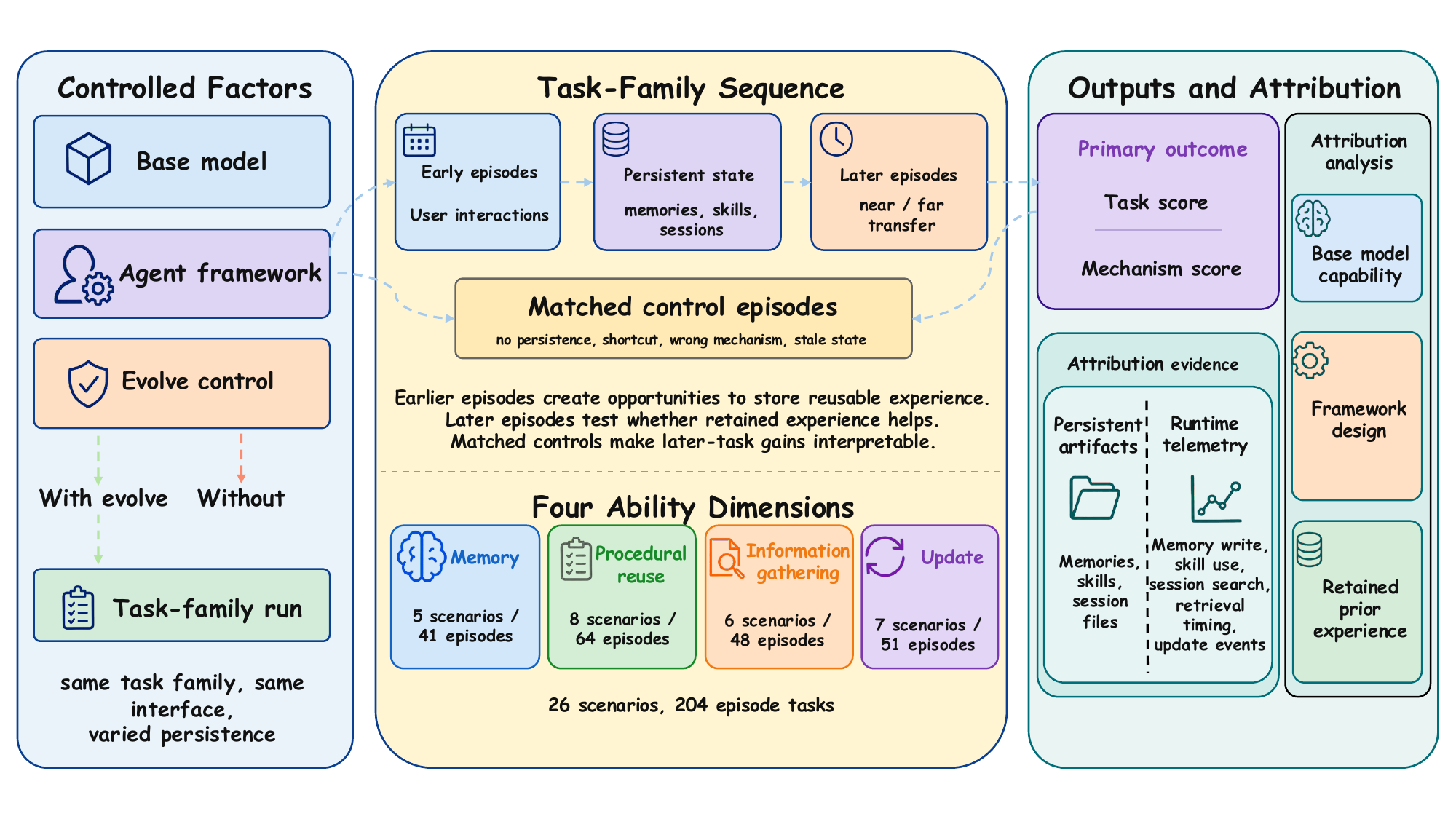}
  \vspace{-0.2in}
  \caption{Overview of \benchname. The benchmark tests whether agents improve across sessions by
reusing retained experience, covering four capability dimensions, 26 task-family scenarios,
and 204 episode tasks with matched no-persistence controls.}
  \label{fig:benchmark_overview}
\end{figure}

\newpage
\vspace{0.5em}
{
  \hypersetup{linkcolor=black}
  \setlength{\parskip}{0pt}
  \renewcommand{\contentsname}{\normalfont\large\bfseries Contents}
  \setcounter{tocdepth}{3}
  \begingroup
    \small
    \tableofcontents
  \endgroup
}
\newpage
\pagestyle{fancy}

\input{intro}
\input{main_related_work}

\input{main_benchmark}
\section{Experiments}
\label{sec:experiments_results_analysis}

We first describe the controlled self-evolution setting (Section~\ref{subsec:setup}), report main results across models and agent frameworks (Section~\ref{subsec:main_results}), use those results to diagnose where self-evolution breaks down and introduce \ouragent (Section~\ref{subsec:hermes_plus_design}), and finally check whether the diagnosed fixes show up in mechanism ablations and on other base models (Section~\ref{subsec:ablation_evidence}).

\subsection{Experimental Setup}
\label{subsec:setup}

Each \benchname task family is evaluated under \textbf{paired persistence conditions}: persistence-off (no retained state from the learning episode) and persistence-on (the agent may reuse memory, skills, profile state, or session history). We report the persistence-on score and the family-balanced gap $\Delta$. This gap measures whether retained experience improves later-task performance. Per-episode computational costs (tokens, wall time) are reported in Table~\ref{tab:computational_cost} in Appendix~\ref{app:computational_cost}.

$\Delta$ alone is not sufficient evidence of self-evolution. We also report a \textbf{mechanism-evidence score} based on saved artifacts and runtime telemetry (see the Mechanism-Evidence Score subsection and Equation~\ref{eq:mech_family} in Appendix~\ref{app:metric_details}), asking whether the improvement used the intended persistence pathway. Throughout this section, task score and $\Delta$ measure behavior; mechanism evidence supports attribution.

\subsection{Main Results}
\label{subsec:main_results}

\begin{table}[t]
  \caption{\textbf{Overall performance of \benchname with Hermes agent.} w/o evolve and w/ evolve denote persistence off and on; $\Delta$ is their family-balanced difference. Results are reported in average across 3 runs. Subscripts on capability $\Delta$ cells give each capability's signed share of the row's total absolute movement $\sum_c |\Delta_c|$ (per-row absolute values sum to 100\%); the Overall $\Delta$ column is the macro-average and carries no subscript. Mech.\ is the
  mechanism-evidence score when applicable.
  }
  \label{tab:main_results_by_ability}
  \centering
  \footnotesize
  \setlength{\tabcolsep}{2.8pt}
  \renewcommand{\arraystretch}{1.08}
  \begin{tabular}{llcccccc}
    \toprule
    Model & State & Memory & Procedural & Info. & Update & Overall & Mech. \\
    \midrule
    GLM-5.1 & w/o evolve & $0.51$ & $0.44$ & $0.59$ & $0.53$ & $0.52$ & \benchdash \\
    & w/ evolve & $0.74$ & $0.53$ & $0.69$ & $0.90$ & $0.71$ & $0.70$ \\
    & $\Delta$ & $+0.23\dpct{29}$ & $+0.09\dpct{11}$ & $\mathbf{+0.11}\dpct{14}$ & $\mathbf{+0.36}\dpct{46}$ & $\underline{+0.20}$ & \benchdash \\
    \midrule
    Kimi K2.6 & w/o evolve & $0.51$ & $0.60$ & $0.65$ & $0.50$ & $0.57$ & \benchdash \\
    & w/ evolve & $0.83$ & $0.63$ & $0.70$ & $0.78$ & $0.73$ & $0.72$ \\
    & $\Delta$ & $\underline{+0.33}\dpct{49}$ & $+0.03\dpct{4}$ & $+0.05\dpct{7}$ & $+0.27\dpct{40}$ & $+0.17$ & \benchdash \\
    \midrule
    DeepSeek-V4-Pro & w/o evolve & $0.46$ & $0.42$ & $0.60$ & $0.51$ & $0.50$ & \benchdash \\
    & w/ evolve & $0.79$ & $0.54$ & $0.66$ & $0.69$ & $0.67$ & $0.71$ \\
    & $\Delta$ & $\underline{+0.33}\dpct{48}$ & $+0.12\dpct{17}$ & $+0.06\dpct{9}$ & $+0.18\dpct{26}$ & $+0.17$ & \benchdash \\
    \midrule
    MiniMax-M2.7 & w/o evolve & $0.50$ & $0.50$ & $0.62$ & $0.50$ & $0.53$ & \benchdash \\
    & w/ evolve & $0.77$ & $0.55$ & $0.71$ & $0.62$ & $0.66$ & $0.64$ \\
    & $\Delta$ & $+0.26\dpct{50}$ & $+0.05\dpct{10}$ & $+0.09\dpct{17}$ & $+0.12\dpct{23}$ & $+0.13$ & \benchdash \\
    \midrule
    GPT-5.4 & w/o evolve & $0.48$ & $0.49$ & $0.62$ & $0.46$ & $0.51$ & \benchdash \\
    & w/ evolve & $0.85$ & $0.68$ & $0.69$ & $0.80$ & $0.75$ & $\mathbf{0.80}$ \\
    & $\Delta$ & $\mathbf{+0.37}\dpct{38}$ & $\mathbf{+0.19}\dpct{20}$  & $+0.07\dpct{7}$ & $\underline{+0.34}\dpct{35}$ & $\mathbf{+0.24}$ & \benchdash \\
    \midrule
    Claude Sonnet 4.6 & w/o evolve & $0.51$ & $0.43$ & $0.59$ & $0.48$ & $0.50$ & \benchdash \\
    & w/ evolve & $0.83$ & $0.53$ & $0.69$& $0.77$ & $0.71$ & $\underline{0.76}$ \\
    & $\Delta$ & $+0.32\dpct{40}$ & $+0.10\dpct{12}$ & $\underline{+0.10}\dpct{12}$ & $+0.29\dpct{36}$ & $\underline{+0.20}$ & \benchdash \\
    \midrule
     Claude Opus 4.6 & w/o evolve & $0.32$ & $0.50$ & $0.62$ & $0.50$ & $0.49$ & \benchdash \\
    & w/ evolve & $0.58$ & $0.64$ & $0.68$ & $0.78$ & $0.67$ & $0.70$ \\
    & $\Delta$ & $+0.26\dpct{35}$ & $\underline{+0.14}\dpct{19}$ & $+0.06\dpct{8}$ & $+0.28\dpct{38}$ & $+0.19$ & \benchdash \\
    \bottomrule
  \end{tabular}
\end{table}

Table~\ref{tab:main_results_by_ability} fixes the agent framework to Hermes and varies the base
model. The leftmost column records the evolution state: persistence-off (w/o evolve), persistence-on (w/ evolve), or their difference ($\Delta$). Mechanism score is reported separately so that higher accuracy is not confused with evidence-aligned self-evolution. Table~\ref{tab:agent_results_by_ability} fixes the model to MiniMax-M2.7 and varies the agent framework, separating model effects from runtime effects.

\begin{table}[t]
  \caption{Fixed-model agent comparison on the four \benchname capabilities. MiniMax-M2.7 is held fixed
  and agent frameworks vary. Capability columns report persistence-on/off deltas; Mech.\ summarizes whether improved outcomes align with expected artifacts or telemetry. Subscripts on capability $\Delta$ cells give each capability's signed share of $\sum_c |\Delta_c|$ (per-row absolute values sum to 100\%); the Overall $\Delta$ column is the macro-average and carries no subscript. Missing
  runs are shown as dashes. Sources for Agent-Zero, nanobot, and ZeroClaw are
  \citep{nanobot2026,zeroclaw2026,agentzero2026}.}
  \label{tab:agent_results_by_ability}
  \centering
  \footnotesize
  \setlength{\tabcolsep}{3.2pt}
  \renewcommand{\arraystretch}{1.08}
  \begin{tabular}{lcccccc}
    \toprule
    Agent & Memory $\Delta$ & Procedural $\Delta$ & Info. $\Delta$ & Update $\Delta$ &
    Overall $\Delta$ & Mech. \\
    \midrule
    nanobot & $+0.06\dpct{10}$ & $-0.06\dpct{-10}$ & $\mathbf{+0.13}\dpct{21}$ & $\mathbf{+0.37}\dpct{59}$ & $\underline{+0.13}$ & $0.57$ \\
    ZeroClaw & $\mathbf{+0.29}\dpct{52}$ & $-0.04\dpct{-7}$ & $+0.08\dpct{14}$ & $+0.15\dpct{27}$ & $+0.12$ & $0.55$ \\
    Agent-Zero & $-0.27\dpct{-52}$ & $\mathbf{+0.11}\dpct{21}$ & $-0.01\dpct{-2}$ & $-0.13\dpct{-25}$ & $-0.08$ & $0.39$ \\
    Hermes & $+0.26\dpct{50}$ & $\underline{+0.05}\dpct{10}$ & $+0.09\dpct{17}$ & $+0.12\dpct{23}$ & $\underline{+0.13}$ & $\underline{0.64}$ \\
    \ouragent (Our framework) & $\underline{+0.27}\dpct{42}$ & $-0.02\dpct{-3}$ & $\underline{+0.12}\dpct{18}$ & $\underline{+0.24}\dpct{37}$ & $\mathbf{+0.15}$ & $\mathbf{0.73}$ \\
    \bottomrule
  \end{tabular}
\end{table}

\paragraph{Self-evolving agent frameworks are robust across base models, but the subtasks that benefit most depend on each model's strengths.}
Every base model in Table~\ref{tab:main_results_by_ability} gains from persistence (Overall $\Delta$ from $+0.13$ to $+0.24$), so the runtime carries persistence value across architectures. Where each model concentrates that gain, however, varies sharply with its own profile (capability subscripts on the $\Delta$ rows): GPT-5.4 spreads its movement evenly across Memory ($38\%$) and Update ($35\%$); GLM-5.1 places nearly half of its movement on Update ($46\%$); Kimi K2.6 places nearly half on Memory ($49\%$); DeepSeek-V4-Pro and Claude Sonnet~4.6 sit between these extremes. The capability the model already excels at is also where retained experience helps it most, which is why a single Overall $\Delta$ tells the wrong story: the four-capability decomposition is what reveals the model-specific strengths.

\paragraph{Advanced agent frameworks show diverse strengths across task subtypes.}
Each framework in Table~\ref{tab:agent_results_by_ability} concentrates its movement on a different subtype (capability subscripts): ZeroClaw lifts Memory ($52\%$ of its movement) but loses ground on Procedural; nanobot puts $60\%$ on Update yet barely improves Memory; Agent-Zero regresses on three of four capabilities. Hermes is the only baseline framework that moves all four capabilities upward; \ouragent lifts Memory, Information Gathering, Update, and Overall but shows a small Procedural dip ($-0.02$). Even with this dip, \ouragent shifts more of its gain onto Update ($37\%$ vs.\ Hermes's $23\%$) without sacrificing Memory. The mechanism-evidence score moves with this shape: nanobot and Hermes both reach $\Delta = +0.13$, but nanobot earns it from a single capability with no consistent write-then-read trace, dropping its Mech to $0.57$ against Hermes's $0.64$. The same headline $\Delta$ can hide two completely different ways of getting there (Figure~\ref{fig:agent_attribution_frontier} in Appendix~\ref{app:agent_attribution_frontier} plots the frontier).

\begin{figure}[t]
  \centering
  \includegraphics[width=0.99\linewidth]{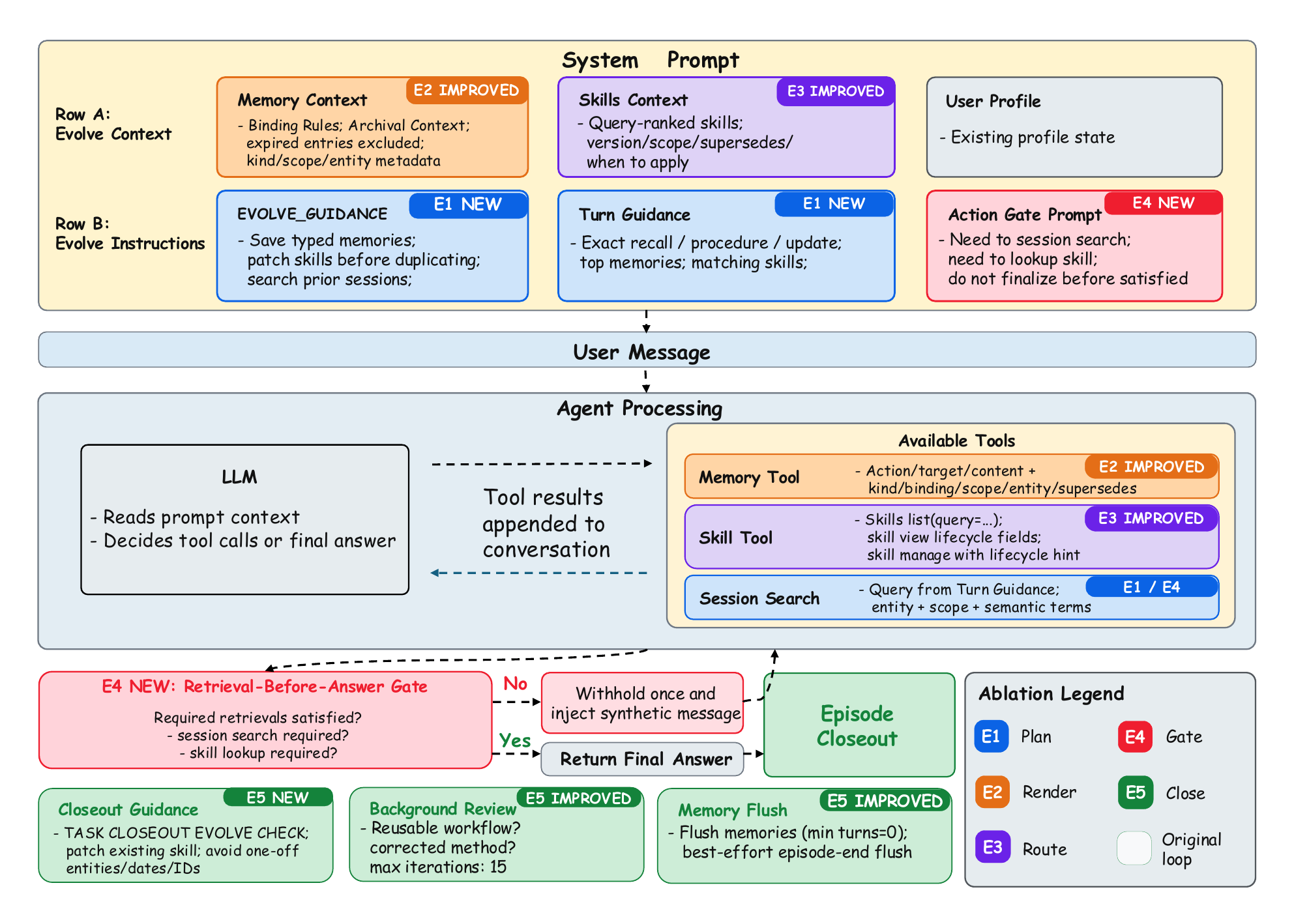}
  \caption{Runtime insertion points in \ouragent. Gray boxes are the original Hermes loop; colored boxes mark the added or modified decisions in prompt context, tools, retrieval gating, and episode closeout. Colors follow the ablation labels: E1 Plan, E2 Render, E3 Route, E4 Gate, and E5 Close.}
  \label{fig:intervention_template}
\end{figure}

\subsection{Diagnosis-Driven Design: \ouragent}
\label{subsec:hermes_plus_design}

Section~\ref{subsec:main_results} uses \benchname as a diagnostic tool. Table~\ref{tab:main_results_by_ability} fixes the framework and varies the model, showing that different models gain on different capabilities. Table~\ref{tab:agent_results_by_ability} fixes the model and varies the framework, showing that the same task-score gain can come with different mechanism evidence. We then inspected low-gain and uneven-gain traces. The failures were concrete and fell into five disjoint categories: plans were drafted without first consulting saved state, saved facts appeared in the wrong form, learned procedures stayed outside the skill library, stored evidence was skipped before action, and corrected state failed to reach the next fresh session. Each category is the responsibility of a single loop stage, and we treat each as the responsibility of one mechanism so that the mechanisms can be enabled or ablated independently of one another.

We choose Hermes for the intervention study because it is the strongest controlled baseline among the existing agents. In Table~\ref{tab:agent_results_by_ability}, Hermes is the only non-\ouragent framework with positive $\Delta$ on all four capabilities. It also ties nanobot for the best baseline Overall $\Delta$ ($+0.13$) while showing stronger mechanism evidence ($0.64$ vs.\ $0.57$); ZeroClaw and Agent-Zero are lower on Mech ($0.55$ and $0.39$). Hermes already exposes the persistence surfaces \benchname probes: memory records, user-profile state, skills, and session history. This lets us keep the model, tasks, grader, and substrate fixed while changing only the runtime decisions. Here, runtime means the decisions that choose when to read, write, or trigger the persistence surfaces.

\textbf{The plan does not condition on saved state (cross-cutting).} A failure that recurs across all four capabilities: even when the relevant state has been correctly written and is retrievable, the agent's plan is often drafted without first consulting it, so a draft action gets committed before any saved binding, skill, or rule is brought into the planning context. This is upstream of the ability-specific failures below, since an unconditioned plan can override correctly stored state regardless of how that state is shaped. \textbf{We add a plan-time consultation check (E1).} E1 sits at the planning stage of the loop and, before drafting any risky or recall-dependent action, requires the agent to consult whichever persistent state the runtime currently exposes and to condition the plan on it. E1 reads the typed schema produced by E2 when E2 is active and reads the native Hermes memory records otherwise, so it operates as a stand-alone plan-time gate that does not require any of the other four mechanisms to be enabled.

\textbf{The saved memory is hard to reuse in a fresh session (Memory).} Memory families test declarative facts such as preferences, constraints, and corrected values. Hermes can store these facts, but the storage design is too loose: current and stale notes can sit together, scope is implicit, and the next session may not see a clear valid clause to apply. \textbf{We store memory as typed bindings and render only the valid one (E2).} E2 writes each memory with type, scope, entity, current value, superseded value, and expiry, and at render time surfaces only the in-scope current binding to the next session, suppressing the superseded entries that would otherwise compete with it.

\textbf{The solved workflow is not saved as an executable procedure (Procedural).} Procedural families test ordered execution, not one-shot fact lookup: the agent must reopen a workflow and follow its steps. Hermes traces show successful learn episodes whose SOP remains in transcript text or splits into near-duplicate notes. Later episodes ask for the saved procedure, but there is no ranked skill to open and execute. \textbf{We save procedures as ranked, patchable skills (E3).} E3 writes a skill with an applicability condition and ordered steps, ranks saved skills by query relevance, and directs the model to patch the closest existing skill when the workflow changes.

\textbf{The agent acts before checking stored evidence (Information Gathering).} Information-gathering families preseed the needed evidence into memory or session history. The failure is trigger timing: under noisy prompts, Hermes may answer from visible context, refuse to guess, or take an irreversible action before calling the persistence channel. \textbf{We require retrieval before recall-dependent actions (E4).} E4 blocks a draft answer when the task depends on prior state and no persistence read has occurred, then requires a read from the relevant channel.

\textbf{Old persistent evidence remains active after correction (Update).} Update families provide a second authoritative value and then test whether the agent uses it in a fresh session. Hermes traces often record the correction inside the current session, while the next episode reads an older artifact or an unstructured transcript fragment. The result is a stale answer after the user has already corrected it. \textbf{We make the new persistent evidence overwrite the old one (E5).} E5 extracts the final binding key or updated rule at episode close, writes it as the new authoritative artifact in place of the prior value, and flushes it synchronously to the persistent store, so the next fresh session retrieves only the corrected value rather than reading the older artifact alongside it.

Together, the cross-cutting plan-time check and the four ability-specific fixes map one-to-one onto the five colored runtime insertion points in Figure~\ref{fig:intervention_template}. Each mechanism is wired as an independent drop-in at its own loop stage, with the remaining stages held at the Hermes default whenever a mechanism is run in isolation. Figures~\ref{fig:case_plan}--\ref{fig:case_close} in Appendix~\ref{app:mechanism_case_studies} give trace-backed case studies for the corresponding mechanisms.

\subsection{Mechanism Evidence and Generalization}
\label{subsec:ablation_evidence}

Table~\ref{tab:ablation_by_ability} isolates the contribution of each mechanism. Each non-final row adds a single mechanism on top of Base Hermes, leaving all other loop stages at their defaults; rows are therefore \emph{not} cumulative. The final row, \ouragent~(full), turns on all five mechanisms simultaneously. All settings share a fixed model (MiniMax-M2.7), task set, and grader. For each capability we report two numbers: the persistence-on score (w/) and the persistence-on/off gap ($\Delta$).
\begin{table}[t]
\caption{Single-mechanism ablations on \benchname (MiniMax-M2.7, fixed task set and grader). \textbf{Each non-final row adds one mechanism to Base Hermes; rows are \emph{not} cumulative.} The bottom row, \ouragent~(full), turns on all five. Per capability: w/ is the persistence-on score, $\Delta$ the family-balanced persistence-on/off gap. Subscripts give each capability's signed share of $\sum_c |\Delta_c|$ (per-row absolute values sum to 100\%); Overall $\Delta$ is the macro-average and carries no subscript. Best values are \textbf{bold}; second-best, \underline{underlined}.}
  \label{tab:ablation_by_ability}
  \centering
  \scriptsize
  \setlength{\tabcolsep}{1.8pt}
  \renewcommand{\arraystretch}{1.14}
  \begin{tabular}{p{0.21\linewidth}cccccccccc}
    \toprule
    & \multicolumn{2}{c}{Memory} & \multicolumn{2}{c}{Procedural} & \multicolumn{2}{c}{Info.} & \multicolumn{2}{c}{Update} & \multicolumn{2}{c}{Overall} \\
    \cmidrule(lr){2-3} \cmidrule(lr){4-5} \cmidrule(lr){6-7} \cmidrule(lr){8-9} \cmidrule(lr){10-11}
    Setting & w/ & $\Delta$ & w/ & $\Delta$ & w/ & $\Delta$ & w/ & $\Delta$ & w/ & $\Delta$ \\
    \midrule
    Base Hermes & $0.77$ & $+0.26\dpct{50}$ & $\mathbf{0.55}$ & $+0.05\dpct{10}$ & $0.71$ & $+0.09\dpct{17}$ & $0.62$ & $+0.12\dpct{23}$ & $\mathbf{0.66}$ & $+0.13$ \\
    + planning guidance (E1) & $0.73$ & $\underline{+0.32}\dpct{58}$ & $0.44$ & $+0.05\dpct{9}$ & $0.68$ & $+0.05\dpct{9}$ & $0.62$ & $+0.13\dpct{24}$ & $\underline{0.62}$ & $+0.14$ \\
    + memory binding (E2) & $\mathbf{0.80}$ & $+0.30\dpct{54}$ & $0.30$ & $-0.02\dpct{-4}$ & $0.71$ & $+0.14\dpct{25}$ & $0.57$ & $+0.10\dpct{18}$ & $0.60$ & $+0.13$ \\
    + skill lifecycle (E3) & $0.46$ & $+0.17\dpct{36}$ & $\underline{0.47}$ & $\mathbf{+0.10}\dpct{21}$ & $\underline{0.75}$ & $\underline{+0.15}\dpct{32}$ & $0.35$ & $+0.05\dpct{11}$ & $0.51$ & $+0.12$ \\
    + retrieval gate (E4) & $0.63$ & $\mathbf{+0.36}\dpct{54}$ & $0.43$ & $\underline{+0.08}\dpct{12}$ & $\mathbf{0.78}$ & $\mathbf{+0.17}\dpct{25}$ & $0.37$ & $+0.06\dpct{9}$ & $0.55$ & $\mathbf{+0.17}$ \\
    + closeout / flush (E5) & $0.49$ & $+0.20\dpct{42}$ & $0.33$ & $+0.00\dpct{0}$ & $0.69$ & $+0.12\dpct{25}$ & $\underline{0.70}$ & $\underline{+0.16}\dpct{33}$ & $0.55$ & $+0.12$ \\
    \ouragent (full) & $\underline{0.78}$ & $+0.27\dpct{42}$ & $0.38$ & $-0.02\dpct{-3}$ & $0.73$ & $+0.12\dpct{18}$ & $\mathbf{0.74}$ & $\mathbf{+0.24}\dpct{37}$ & $\mathbf{0.66}$ & $\underline{+0.15}$ \\
    \bottomrule
  \end{tabular}
\end{table}

\begingroup
\setlength{\parskip}{0.2\baselineskip}

\paragraph{Single mechanisms support their target diagnoses.} E2 (Render) gives the highest Memory persistence-on score ($0.80$), E3 (Route) the largest single-mechanism Procedural $\Delta$ ($+0.10$), E4 (Gate) the largest Info $\Delta$ ($+0.17$), and E5 (Close) the strongest single-mechanism Update $\Delta$ ($+0.16$). Together, these results align with the failure-to-mechanism mapping in Section~\ref{subsec:hermes_plus_design}.

\paragraph{The full \ouragent preserves overall task performance and has its clearest gain on Update.} \ouragent (full) ties Base Hermes on Overall persistence-on score ($0.66$), raises the reported Overall $\Delta$ from $+0.13$ to $+0.15$, and reaches the best Update score ($0.74$) and gap ($+0.24$), while its Procedural result declines slightly. Figure~\ref{fig:ablation_heatmap} provides the per-capability view.

\Needspace{19\baselineskip}
\begin{wraptable}{r}{0.49\linewidth}
  \centering
  \caption{Focused Procedural interaction diagnosis.}
  \label{tab:procedural_interaction_compact}
  \scriptsize
  \setlength{\tabcolsep}{2.0pt}
  \renewcommand{\arraystretch}{1.02}
  \begin{tabularx}{\linewidth}{@{}Xcc@{}}
    \toprule
    \multicolumn{3}{@{}l}{\textit{Full-minus-one persistence gaps}} \\
    Setting & \multicolumn{2}{c}{$\Delta$} \\
    \midrule
    Base Hermes & \multicolumn{2}{c}{$\underline{+0.087}$} \\
    Full \ouragent & \multicolumn{2}{c}{$+0.085$} \\
    w/o E1 & \multicolumn{2}{c}{$+0.084$} \\
    w/o E2 & \multicolumn{2}{c}{$\mathbf{+0.108}$} \\
    w/o E3 & \multicolumn{2}{c}{$+0.062$} \\
    w/o E4 & \multicolumn{2}{c}{$+0.082$} \\
    w/o E5 & \multicolumn{2}{c}{$+0.042$} \\
    \midrule
    \multicolumn{3}{@{}l}{\textit{Routing comparison}} \\
    Measure & Full & w/o E2 \\
    \midrule
    No task-specific skill before evaluation & 2/6 & 1/6 \\
    No task-specific skill read during evaluation & 3/6 & 2/6 \\
    \bottomrule
  \end{tabularx}
\end{wraptable}

\paragraph{A focused Procedural diagnosis exposes mechanism interaction.} The full-minus-one rows in Table~\ref{tab:procedural_interaction_compact} show that removing E2 raises $\Delta$ from $+0.085$ to $+0.108$, whereas removing E3 or E5 lowers it to $+0.062$ or $+0.042$. The routing rows link the E2 effect to more consistent task-specific skill creation and reuse. We treat this as a focused diagnostic rather than a full-benchmark estimate; Appendix~\ref{app:procedural_diagnosis} provides a representative trace.

\paragraph{\ouragent improves or preserves the Hermes baseline on most base models.} Swapping \ouragent (tuned on MiniMax-M2.7) onto five base models matches or improves each model's Hermes baseline on three of five configurations --- MiniMax-M2.7 ($+0.13 \to +0.15$), Claude Sonnet 4.6 ($+0.20 \to +0.22$), and GPT-5.4 (flat at $+0.24$). On its strongest pairing, \ouragent + GPT-5.4 ties the benchmark's highest configuration ($\Delta = +0.24$, Mech $0.80$). DeepSeek-V4-Pro and Claude Opus 4.6 regress slightly, so the transfer result is positive but not uniform. Table~\ref{tab:selected_model_scores} gives the full per-capability comparison.

\paragraph{The three analyses answer different attribution questions.} Table~\ref{tab:ablation_by_ability} tests isolated interventions, Table~\ref{tab:procedural_interaction_compact} tests mechanism interactions, and Table~\ref{tab:selected_model_scores} tests cross-model transfer. Together they show why Overall $\Delta$ alone is insufficient: target gains can coexist with regressions elsewhere, component effects can reverse in combination, and a runtime change need not transfer uniformly.

\begin{table}[t]
  \caption{\ouragent across five base models. For each capability, w/ is the persistence-on score and $\Delta$ is the family-balanced persistence-on/off gap. Subscripts on capability $\Delta$ cells give each capability's signed share of $\sum_c |\Delta_c|$ (per-row absolute values sum to 100\%); the Overall $\Delta$ column is the macro-average and carries no subscript. Mech.\ reports mechanism-evidence alignment.}
  \label{tab:selected_model_scores}
  \centering
  \scriptsize
  \setlength{\tabcolsep}{1.8pt}
  \renewcommand{\arraystretch}{1.14}
  \begin{tabular}{p{0.15\linewidth}ccccccccccc}
    \toprule
    & \multicolumn{2}{c}{Memory} & \multicolumn{2}{c}{Procedural} & \multicolumn{2}{c}{Info.} & \multicolumn{2}{c}{Update} & \multicolumn{2}{c}{Overall} & Mech. \\
    \cmidrule(lr){2-3} \cmidrule(lr){4-5} \cmidrule(lr){6-7} \cmidrule(lr){8-9} \cmidrule(lr){10-11}
    Model & w/ & $\Delta$ & w/ & $\Delta$ & w/ & $\Delta$ & w/ & $\Delta$ & w/ & $\Delta$ & Score \\
    \midrule
    DeepSeek-V4-Pro & $\underline{0.87}$ & $\mathbf{+0.42}\dpct{70}$ & $0.49$ & $+0.04\dpct{7}$ & $0.62$ & $+0.01\dpct{2}$ & $0.49$ & $+0.13\dpct{22}$ & $0.62$ & $+0.15$ & $0.69$ \\
    MiniMax-M2.7 & $0.78$ & $+0.27\dpct{42}$ & $0.38$ & $-0.02\dpct{-3}$ & $\mathbf{0.73}$ & $\mathbf{+0.12}\dpct{18}$ & $0.74$ & $+0.24\dpct{37}$ & $0.66$ & $+0.15$ & $0.73$ \\
    GPT-5.4 & $\mathbf{0.88}$ & $\underline{+0.41}\dpct{43}$ & $\underline{0.68}$ & $\mathbf{+0.18}\dpct{19}$ & $\underline{0.70}$ & $+0.06\dpct{6}$ & $0.81$ & $+0.31\dpct{32}$ & $\mathbf{0.77}$ & $\mathbf{+0.24}$ & $\mathbf{0.80}$ \\
    Claude Sonnet 4.6 & $\underline{0.87}$ & $+0.33\dpct{38}$ & $\mathbf{0.69}$ & $\underline{+0.17}\dpct{19}$ & $\underline{0.70}$ & $+0.05\dpct{6}$ & $\underline{0.83}$ & $\mathbf{+0.33}\dpct{38}$ & $\underline{0.77}$ & $\underline{+0.22}$ & $\underline{0.77}$ \\
    Claude Opus 4.6 & $0.68$ & $+0.27\dpct{36}$ & $0.52$ & $+0.07\dpct{10}$ & $0.69$ & $\underline{+0.07}\dpct{10}$ & $\mathbf{0.84}$ & $\mathbf{+0.33}\dpct{45}$ & $0.68$ & $+0.18$ & $0.70$ \\
    \bottomrule
  \end{tabular}
\end{table}

\paragraph{Run-to-run variation tempers the aggregate comparison.} Across three MiniMax-M2.7 runs, the Overall gap is $0.13 \pm 0.04$ for Hermes and $0.15 \pm 0.06$ for \ouragent. The $+0.02$ difference is smaller than the run-to-run variation, so we do not interpret it as a stable overall gain. The clearer mean shift is on Update ($+0.12 \to +0.24$), although its $\sigma_\Delta$ also increases from $0.01$ to $0.09$; Appendix~\ref{app:variance_analysis} (Table~\ref{tab:variance_by_ability}) gives the full capability-level breakdown.

\endgroup

\section{Conclusion}
\label{sec:conclusion}

We introduced \benchname, a performance-attribution benchmark that pairs persistence-on/off evaluations within task families and reports mechanism evidence alongside task scores, separating base-model, runtime, and retained-experience contributions to later-task performance. Experiments across seven models and four frameworks show that self-evolution is capability-specific and that similar persistence gaps can hide different persistence paths. Under MiniMax-M2.7, \ouragent raises the reported mean Overall $\Delta$ from $+0.13$ to $+0.15$ and Mech from $0.64$ to $0.73$, with its clearest gain on Update; the $+0.02$ Overall difference is smaller than run-to-run variation. The effect is not uniform across capabilities or base models, so we treat \ouragent as a diagnostic scaffold rather than a universal improvement.


\section{Future Work}
PAST-Bench provides an initial foundation for attributing cross-session improvement to retained experience, but several directions remain open. First, future versions should broaden the ecological validity and temporal scope of the benchmark. The current task families are synthetically constructed and evaluated in isolation. An important next step is to incorporate human-authored and interaction-derived scenarios, longer task sequences, and settings in which experience accumulated in one family affects behavior in another. Such extensions would test whether persistent agents can maintain useful state over longer horizons, transfer experience across changing domains, and avoid interference among independently acquired memories, procedures, and corrections.

Second, the capability space should be expanded beyond memory, procedural reuse, information gathering, and update. These capabilities represent necessary foundations of online self-evolution, but they do not cover stronger forms of recursive improvement. Future benchmarks could evaluate whether agents acquire previously unavailable tool-use strategies, construct and revise long-horizon plans, coordinate experience across multiple agents, and improve the mechanisms by which they decide what to store, retrieve, verify, and update. This would help distinguish systems that merely reuse retained state from systems that improve their own learning and adaptation processes.

Third, future work should strengthen mechanism attribution. The current mechanism-evidence score measures consistency with an expected persistence pathway, rather than establishing causal necessity. A stronger evaluation could combine trace evidence with counterfactual interventions, such as deleting, replacing, or corrupting a candidate artifact and measuring the resulting behavioral change. It would also be useful to support multiple semantically valid persistence pathways, since different agents may encode the same experience as a memory, skill, structured artifact, or revised policy. Larger-scale human pathway annotations and uncertainty estimates would further improve the construct validity of mechanism-level evaluation.

Finally, the capability-specific and model-dependent behavior observed in Hermes+ suggests that persistence mechanisms should not be treated as uniformly composable. Future agents could learn to route experience dynamically across memory, skills, and session history, while detecting conflicts, redundancy, and stale state across these substrates. In particular, the interaction between structured memory rendering and procedural skill routing motivates adaptive mechanisms that decide not only when to read or write persistent state, but also which persistence surface should own a given piece of experience. Developing such mechanisms under explicit accuracy, latency, and token-cost constraints may provide a practical path from persistent agents that retain experience to agents that systematically improve through it.

\section*{Acknowledgments}
Z Ding and Y Chen are supported by the U.S. National Science Foundation (NSF) under grants 2037026, 2313131, 2543755 and 2607613.

\newpage
\bibliographystyle{abbrvnat}
\bibliography{clawbench_related_work_refs}

\clearpage
\appendix
\input{clawbench_family_examples_appendix}   
\input{clawbench_metric_details_appendix}
\input{agent_frameworks_appendix}            
\input{appendix}                              
\input{reproducibility_appendix}
\input{related_work_appendix}                 

\end{document}

%% file: intro.tex
\section{Introduction}
\label{sec:introduction}

Recursive self-improvement (RSI) concerns the ability of an AI system to use experience generated through its own operation to improve its future capabilities~\citep{wang2026openclaw,ren2026selfimprovements,lee2026recursive,qu2024recursive,yin2025godel}.
While stronger forms of RSI may eventually involve modifying model parameters, learning algorithms, or agent architectures, a more immediate and operational layer is already emerging in personal AI agents~\citep{wang2026openclaw,gao2025survey,sarukkai2025selfgenerated}.
Personal AI agents now persist across sessions. They read messages, operate over user workspaces, call tools, and accumulate files, memories, skills, and session histories over days and months~\citep{openclaw2026,hermes2026}. Agent frameworks such as Hermes~\citep{hermes2026} and OpenClaw~\citep{openclaw2026} treat persistent workspaces, memories, skills, and tool execution as first-class runtime components, while memory-layer systems such as Mem0~\citep{mem02026} and LangGraph~\citep{langgraphdeepagents2026} provide the substrate: editable memory, interaction-derived facts, temporal knowledge graphs, and procedural skill files. In these systems, user interactions are no longer merely transient context; they can become experience that changes the agent's future behavior.

Personal agents thus provide a natural, user-grounded testbed for learning from experience. Before an agent can recursively improve the mechanisms by which it learns, reasons, or acts, it must first close a more basic loop: identifying useful experience, preserving it beyond the current session, retrieving it when relevant, applying it correctly, and revising it when it becomes outdated~\citep{xu2026amem}. This shifts the unit of evaluation. The relevant question is no longer whether an agent solves the current task, but whether it becomes \emph{better} at serving the same user across future ones---retaining durable preferences, reusing prior workflows, and revising stale information~\citep{buening2026aligning}. We call this capability \emph{online self-evolution}: a personal agent changes its future behavior by reusing experience accumulated during prior interactions, without model retraining~\citep{xia2025agent0,ou2025symbolic}, prompt optimization~\citep{khattab2024dspy,yuksekgonul2025optimizing}, or long-context adaptation~\citep{agarwal2024many}. Online self-evolution is not RSI in its full form, but it provides a concrete behavioral and infrastructural substrate on which stronger forms of recursive improvement can be built~\citep{zhang2026memrl,fang2025comprehensive}.

Accumulating experience does not guarantee improvement. An agent may store the wrong evidence, retrieve irrelevant memory, reuse brittle procedures, or apply stale state to a new task. Evaluating self-evolution is thus a \emph{performance-attribution} problem: if later-session performance improves, the gain might come from retained experience---or from the base model, runtime, prompt, retrieval shortcuts, task difficulty, or scoring noise. Current benchmarks cannot make this distinction. Interactive agent benchmarks~\citep{liu2024agentbench,koh2024visualwebarena,drouin2024workarena,xie2024osworld,merrill2026terminalbench,mialon2023gaia,zhang2026clawbench} reduce evaluation to a one-shot per-task score on a fresh session, not a trajectory. Memory and skill benchmarks~\citep{wu2025longmemeval,maharana2024evaluating,li2026skillsbench} test individual ingredients of persistence in isolation, without matched controls that disentangle retained experience from base-model and runtime contributions.

To this end, we introduce \benchname, a performance-attribution benchmark
built around this question (Figure~\ref{fig:benchmark_overview}). The unit
of evaluation is the agent's \textit{trajectory} through a task family
rather than a one-shot per-task score: earlier episodes give the agent an
opportunity to save reusable experience, later episodes test whether it is
reused, and matched control episodes strip persistence so any later-task
gain can be read against a no-persistence baseline. The current suite
contains 26 scenarios and 204 episodes spanning four capabilities---\emph{
memory} (5/41), \emph{procedural reuse} (8/64), \emph{information gathering}
(6/48), and \emph{update} (7/51)---each targeting a distinct demand on
persistent state. Holding model, task family, and evaluation interface fixed
while toggling persistence makes the with/without gap directly comparable;
saved artifacts and execution traces then reveal whether the agent actually
wrote, retrieved, applied, or revised the state it was supposed to.

\paragraph{Contributions.} We address this gap with a benchmark, diagnostic study, and new agent framework.

\textbf{(1) \benchname: a benchmark for self-evolving personal agents (Section~\ref{sec:ourbench}).} A trajectory-level performance-attribution benchmark: 26 scenarios and 204 episodes across four capabilities, with matched persistence-on/off controls and trace-level evidence enabling per-stage diagnosis of where retained experience helps.

\textbf{(2) Diagnosing self-evolution failures (Section~\ref{sec:experiments_results_analysis}).} Across seven models and four agent frameworks, persistence gaps vary sharply by capability, and agents that tie on the task-score gap can still differ substantially in mechanism evidence (e.g., Hermes vs.\ nanobot: $0.64$ vs.\ $0.57$ at the same $\Delta = +0.13$)---a discrepancy invisible to one-shot scoring.

\textbf{(3) \ouragent: a new agent framework baseline (Section~\ref{sec:experiments_results_analysis}).} \ouragent extends Hermes with five runtime mechanisms, one per stage of the agent loop (Plan, Render, Route, Gate, Close), and serves as a new reference baseline that raises the reported means on both axes ($\Delta$: $+0.13\to\mathbf{+0.15}$, Mech: $0.64\to\mathbf{0.73}$), with super-additive composition on Update ($\Delta = +0.24$, well above any single mechanism: closeout alone $+0.16$, retrieval gate alone $+0.06$). The Overall $\Delta$ difference is smaller than run-to-run variation. We modify Hermes rather than other popular agent frameworks
as it is the only framework in this set that exposes the agent
loop without a pre-instantiated persistence stack, which is the
property required for clean mechanism-by-mechanism ablation (Appendix~\ref{appendix:agent_frameworks}). The others are reported as off-the-shelf baselines.

%% file: main_related_work.tex
\section{Related Work}
\label{sec:related_work}
Prior work on agent evaluation falls into three groups, distinguished by the unit at which evaluation occurs. \benchname{} departs from all three by grading an \emph{episode sequence within a task family} rather than a single task instance, asking whether state created in earlier episodes is reused in later ones. Appendix~\ref{appendix:extended_related_work} contains an extended account of related work.
\paragraph{Interactive and trajectory-level evaluation.} Interactive agent benchmarks~\citep{liu2024agentbench,koh2024visualwebarena,drouin2024workarena,xie2024osworld,merrill2026terminalbench} score complete agent stacks on isolated task instances, conflating base model capability, prompting, tool policy, and any retained experience into a single number. Trajectory-grading benchmarks~\citep{ma2024agentboard,he2025traject,li2026atbench} push further by scoring the action sequence \emph{within} a task. \benchname{} grades whether state produced in earlier tasks is reused \emph{across} later tasks of the same family; within-episode trajectory evidence is an input to this attribution, not the outcome metric.
\paragraph{Memory, skill, and architectural mechanisms.} A second line of work evaluates specific persistence mechanisms in isolation: long-horizon conversational memory~\citep{wu2025longmemeval,maharana2024evaluating}, curated or self-generated skills~\citep{li2026skillsbench,yang2026skillopt,ouyang2026skillos}, and architectural choices~\citep{bogavelli2025agentarch}. These works isolate one substrate but do not test whether retained experience improves later \emph{executable} tasks under family-level controls. \benchname{} uses cold, learning, evaluation, and control episodes to localize improvements to a specific persistence decision.
\paragraph{Mechanism attribution under contamination.} Recent critiques caution that benchmark scores can reflect familiarity with benchmark artifacts rather than transferable problem solving~\citep{liang2025swebenchillusion,berkeleyRDI2026brokenbenchmarks}. The concern applies directly to self-evolution evaluation: later-task gains can be real score gains yet not caused by the persistence mechanism a framework claims credit for. \benchname{} separates outcome gains from trace-level mechanism diagnosis and uses matched persistence-on vs.\ persistence-off controls to check whether the gap is robust to controls for task, model, and runtime. Accordingly, cross-session retention, a well-formed tool trajectory, or a self-generated artifact alone is not evidence of beneficial cross-episode reuse.
Table~\ref{tab:benchmark_comparison} situates \benchname{} against representative agent benchmarks along four methodology axes: cross-session retained experience, fixed-framework model comparison, fixed-model framework comparison, and trajectory-level diagnostics beyond one-shot task success. Existing benchmarks cover strict subsets; \benchname{} is the first to support all four jointly, which retained-experience attribution requires.

\begin{table}[t]
  \caption{Comparison with representative benchmarks. \benchname{} is designed for retained-experience attribution: it evaluates longitudinal task families and supports both model-side and framework-side isolation. \benchyes: axis directly supported. \benchno: not supported. \benchpartial: related proxy tested, not the matched retained-experience comparison.}

  \label{tab:benchmark_comparison}
  \centering
  \footnotesize
  \setlength{\tabcolsep}{3pt}
  \renewcommand{\arraystretch}{1.15}
  \begin{tabular}{p{0.25\linewidth}cccc}
    \toprule
    Benchmark & Retained experience & Model comparison & Framework comparison & Trajectory diagnosis  \\
    \midrule
    \textsc{GAIA}~\citep{mialon2023gaia} & \benchno & \benchno & \benchno & \benchno  \\
    \textsc{AgentBench}~\citep{liu2024agentbench} & \benchno & \benchyes & \benchno & \benchno \\
    \textsc{VisualWebArena}~\citep{koh2024visualwebarena} & \benchno & \benchyes & \benchno & \benchno \\
    \textsc{WorkArena}~\citep{drouin2024workarena} & \benchno & \benchyes & \benchno & \benchno \\
    \textsc{OSWorld}~\citep{xie2024osworld} & \benchno & \benchyes & \benchno & \benchno \\
    \textsc{LongMemEval}~\citep{wu2025longmemeval} & \benchyes & \benchyes & \benchno &
    \benchno \\
    \textsc{LoCoMo}~\citep{maharana2024evaluating} & \benchyes & \benchyes & \benchno &
    \benchno  \\
    \textsc{SkillsBench}~\citep{li2026skillsbench} & \benchpartial & \benchyes &
    \benchpartial & \benchpartial \\
    \textsc{AgentBoard}~\citep{ma2024agentboard} & \benchno & \benchyes & \benchno & \benchyes  \\
    \benchname & \benchyes & \benchyes & \benchyes & \benchyes \\
    \bottomrule
  \end{tabular}
\end{table}

%% file: main_benchmark.tex
\section{PAST-Bench}
\label{sec:ourbench}
Current persistence-aware agent benchmarks fall into one of two regimes. The first keeps prior content visible to the model, either in a long context window~\citep{maharana2024evaluating, letta2025benchmarking} or by incremental injection into a growing dialogue history~\citep{hu2025evaluating}. The second runs sequential task streams in which state propagates across tasks without interruption~\citep{wei2025evo, zheng2025lifelongagentbench}. Both regimes conflate \textit{persistent} learning with \textit{in-context} propagation. \benchname{} instead evaluates online self-evolution under a strict context-clearing protocol. Its evaluation unit is a \emph{task family}: an ordered sequence of fresh-session episodes that share a latent rule (memory), a reusable artifact (procedural reuse), a correction (update), or a pre-seeded reference (information gathering), with the framework's volatile context wiped between episodes. Any improvement of a later episode over an earlier one must therefore flow through the persistent substrate---memory store, skill file, playbook, or edited rule---not through residual prompt overlap.


\subsection{Benchmark Construction}
\label{subsec:benchmark_suite}

\begin{wrapfigure}[18]{r}{0.39\linewidth}
  \centering
  \vspace{-\intextsep}  
  \includegraphics[width=\linewidth]{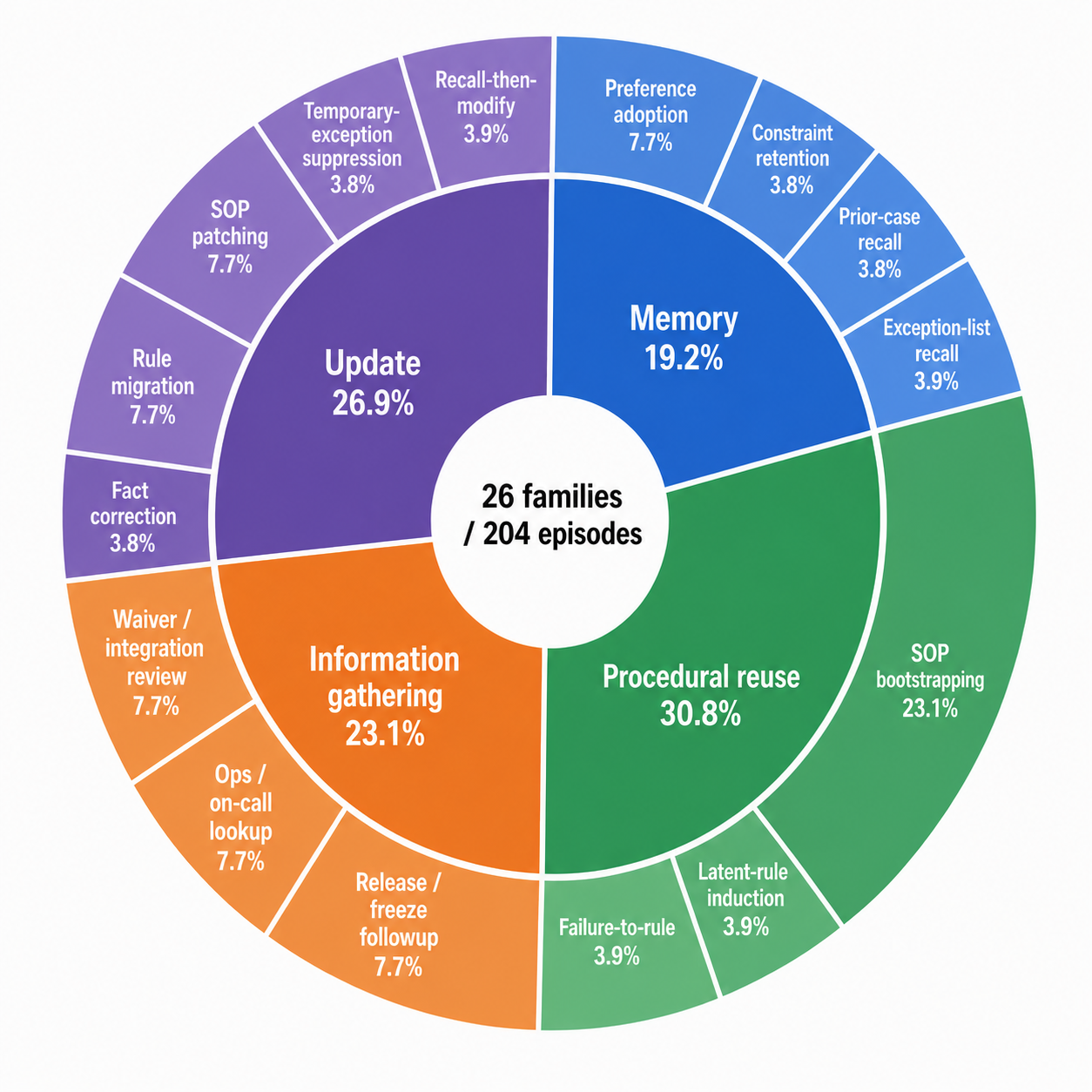}
  \caption{\textbf{Task family distribution of \benchname} across the four capability categories.}
  \label{fig:clawbench_suite_distribution}
\end{wrapfigure}

\benchname targets four
core \textbf{capabilities} of online self-evolution, each requiring retention across sessions and active reuse in later ones: \textbf{Memory}, \textbf{Procedural Reuse}, \textbf{Information Gathering}, and \textbf{Update}.
For each capability, we curate a diverse set of carefully designed \emph{task
families}. Each task family is an ordered sequence of fresh-session episodes sharing a latent rule (\textbf{memory}), a reusable artifact (\textbf{procedural reuse}), a correction (\textbf{update}), or a pre-established reference whose retrieval must be triggered at the right moment (\textbf{information gathering}). For the first three capabilities, earlier episodes expose information, a procedure, or an updated value to be retained; for information gathering, the reference is preseeded before the family begins. In every case, later episodes test whether the agent reuses or consults the relevant state without restating the decisive rule. Each family also contains control episodes: \emph{no-retention} controls that remove the earlier state, \emph{distractor} controls that inject irrelevant or superficially similar information, \emph{stale} controls that expose obsolete memories, and \emph{wrong-mechanism} controls that surface incorrect skills or the wrong evidence source. The distribution of families across capabilities and scenario groups is shown in
Fig.~\ref{fig:clawbench_suite_distribution}. The per-capability breakdown is reported in Table~\ref{tab:clawbench_suite_coverage} in Appendix~\ref{app:family_taxonomy}. Appendix~\ref{app:mechanism_case_studies} presents detailed trace-backed examples in Figures~\ref{fig:case_plan}--\ref{fig:case_close}.

\textbf{Capability 1: General-knowledge memory---can the framework retain and look up an everyday clause?} Memory families isolate the \emph{declarative} pathway~\citep{sumers2023cognitive, squire1992declarative} for casual, user-facing facts (preferences, constraints, one-line policies, exceptions, prior-case decisions). Each family seeds a single read-mostly clause and never revises it; evaluate episodes succeed iff the agent recovers and applies it as a \emph{one-shot lookup-and-apply}, with the trigger wording removed from the prompt. \emph{When} to query is deferred to Capability~3; \emph{revising} a clause is deferred to Capability~4.

\textbf{Capability 2: Professional procedural reuse---can the framework retain and re-execute a multi-step technical workflow?} Procedural-reuse families isolate the \emph{imperative} pathway~\citep{sumers2023cognitive, anderson2014atomic} for domain-specific, technical routines---SOPs, playbooks, build/deploy pipelines, incident-triage flows, induced engineering workflows~\citep{hong2023metagpt}. Where Capability~1 tests \emph{value lookup}, Capability~2 tests \emph{ordered execution} with the right tool composition; order errors, skipped steps, and wrong-tool substitutions are graded as failures. Procedure \emph{revision} is deferred to Capability~4, so this capability isolates first-time procedure formulation.

\textbf{Capability 3: Information gathering---given that the answer is already in the substrate, does the framework consult it at the right moment?} The relevant artifact is \emph{pre-seeded} into the runtime's persistence layer (long-term memory, registered skills, indexed transcripts, or home-state fixtures) before the family begins. The test is not retention but whether the agent \emph{proactively retrieves} under noisy context; each family also plants a generic default that yields a plausible-but-wrong answer if used in place of retrieval.

\textbf{Capability 4: Update---can a second write override a first one without leaking the first?} Update families begin with an existing entry (stale fact, old rule, outdated SOP, or temporary exception) and then deliver an authoritative second write; evaluate episodes score whether the new state is used \emph{and} the old state does not leak. Seven families cover fact correction, global and scoped rule migration, temporary-exception expiry, incident and error-correction SOP patching, and recall-then-modify of a prior session artifact. Figures~\ref{fig:case_plan}--\ref{fig:case_close} in Appendix~\ref{app:mechanism_case_studies} give one representative example per capability.

\subsection{Evaluation Pipeline} 
\label{subsec:evaluation_protocol}
 \paragraph{Episode roles.}
   Every task family is an ordered sequence of fresh-session episodes playing one of four roles. \textbf{Cold} episodes measure first-contact behavior before any persistence can exist. \textbf{Learn} episodes (and, in Update families, an \textbf{Update} episode) deposit the target (a clause, procedure, or correction) into the persistence substrate. \textbf{Evaluation} episodes probe reuse of that state in a later fresh session with the trigger wording removed. \textbf{Control} episodes check that any gain cannot be explained by prompt shortcuts, surface-memorization, stale reuse, or writes to the wrong substrate.  

\paragraph{Persistence and the matched ablation.} By \emph{persistence} we mean benchmark-managed access to state produced or modified by earlier episodes in the same family: memory records, skills, profile entries, session-history indices, saved artifacts, and home-state fixtures. Each evaluation episode is graded under a matched ablation. The \textbf{w/o-evolve} condition denies the runtime any access to family-produced state; the \textbf{w/-evolve} condition permits it. The two runs share the same prompt, grader, tool stack, and seed, so any score gap is most plausibly attributable to the persistence layer rather than to model variance or task drift. We treat this as a strong design control rather than a causal proof; the mechanism-evidence score below provides a complementary substrate-level check. Cold scores are reported for calibration and headroom analysis but are \emph{not} the persistence-off baseline; the matched w/o-evolve condition is.

\paragraph{Reported quantities.} The primary metric for each family $f$ is the \emph{self-evolution gap} $\Delta_f = S^{\text{w/-evolve}}_f - S^{\text{w/o-evolve}}_f$, where $S_f$ is the within-family mean of the per-episode \textbf{task score} $s_e \in [0,1]$. The capability-level $\Delta$ is the macro-average of $\Delta_f$ over families. We accept $\Delta_f$ as evidence of self-evolution only when it clears the upper bound set by the family's control episodes for shortcut, surface-memorization, stale-reuse, and wrong-mechanism explanations. Alongside the gap, \benchname{} reports \textbf{mechanism evidence}: memory writes and reads, skill creation and patching, session-search calls, artifact diffs, and update-correctness signals. An agent that writes artifacts but never reads them, or succeeds through the wrong substrate, should not earn credit on $\Delta$ alone. The task-score definition (Equation~\ref{eq:task_score}) and mechanism-evidence aggregation (Equation~\ref{eq:mech_family}), together with control bounds and per-family rubrics, appear in Appendix~\ref{app:metric_details}.

%% file: clawbench_family_examples_appendix.tex

\section{PAST-Bench Benchmark Details}
\label{app:family_examples}

\subsection{Family Taxonomy}
\label{app:family_taxonomy}

Table~\ref{tab:clawbench_suite_coverage} reports the full task distribution of \benchname.

\begin{table}[ht]
  \caption{The number of task instances in each main capability of task family in \benchname. Each main family contains multiple sub-families covering distinct focus areas. A sample belongs to exactly one main family and exactly one sub-family.}
  \label{tab:clawbench_suite_coverage}
  \centering
  \small
  \begin{tabular}{lclc}
  \toprule
  \textbf{Main Family} & \textbf{\#Episodes} & \textbf{Sub-Family (Focus Area)} & \textbf{\#Episodes} \\
  \midrule
  \multirow{5}{*}{Memory}
    & \multirow{5}{*}{41}
    & Preference Adoption                  & 8 \\
    & & Constraint Retention               & 8 \\
    & & Weak-Trigger Preference Adoption   & 9 \\
    & & Prior Case Recall                  & 8 \\
    & & Exception List Recall              & 8 \\
  \midrule
  \multirow{6}{*}{Information Gathering}
    & \multirow{6}{*}{48}
    & Release Decision Followup            & 8 \\
    & & Ops Exception Desk                 & 8 \\
    & & Oncall Handoff Lookup              & 8 \\
    & & Temporary Waiver Audit             & 8 \\
    & & Change Freeze Followup             & 8 \\
    & & Kappa Integration Review           & 8 \\
  \midrule
  \multirow{8}{*}{Procedural}
    & \multirow{8}{*}{64}
    & SOP Bootstrap 01                 & 8 \\
    & & SOP Bootstrap 02               & 8 \\
    & & SOP Bootstrap 03               & 8 \\
    & & SOP Bootstrap 04               & 8 \\
    & & SOP Bootstrap 05               & 8 \\
    & & SOP Bootstrap 06               & 8 \\
    & & Latent Rule Induction          & 7 \\
    & & Failure-to-Rule                & 9 \\
  \midrule
  \multirow{7}{*}{Update}
    & \multirow{7}{*}{51}
    & Fact Correction                      & 8 \\
    & & Rule Migration                     & 8 \\
    & & Temporary Exception Pollution      & 7 \\
    & & Scoped Rule Migration              & 7 \\
    & & SOP Patch 01                       & 7 \\
    & & SOP Patch 02                       & 7 \\
    & & Recall-then-Modify                 & 7 \\
  \midrule
  \textbf{Total} & \textbf{204} & & \\
  \bottomrule
  \end{tabular}
\end{table}

\subsection{Task Construction and Quality Control}
\label{app:task_construction}

All 26 task families and 204 episodes are synthetic. No task contains data from real users. We first defined the four capabilities from common problems in human--agent interaction: keeping user-specific facts, reusing learned procedures, finding earlier evidence in noisy histories, and replacing outdated state. We then wrote rules for weak triggers, ambiguity, distractors, conflicting records, and transfer across fresh sessions.

The concrete families and episodes were generated from these rules with two model--agent pairs: Codex with GPT-5.4 and Claude Code with Claude Opus~4.6. The prompts, graders, and expected artifacts were generated in the same process. Existing benchmarks informed the high-level scenario taxonomy, but no task was copied from or adapted from another benchmark.

\begin{samepage}
Three authors ran and checked the generated tasks. Each family was checked by at least one author. The check covered six points:
\begin{enumerate}[leftmargin=*,itemsep=1pt,topsep=3pt]
  \item the family and episode roles match the target capability;
  \item each prompt is complete and logically consistent;
  \item ambiguity and weak triggers are intentional;
  \item preferences and corrections stay consistent across episodes;
  \item distractors and controls do not reveal the retained item or create a shortcut; and
  \item graders and expected artifacts match the prompt, partial-credit rules, and stale-answer rules.
\end{enumerate}
\end{samepage}

\subsection{Trace-Backed Mechanism Case Studies}
\label{app:mechanism_case_studies}

Each \benchname family probes a distinct cross-episode dependency that a learn episode establishes and a later eval episode must recover: Constraint Retention (B1), Fact Correction (B2), SOP Bootstrap (B3), Exception List Recall (B4), and Failure-to-Rule (B5) generated by MiniMax-M2.7 under Hermes Agent. Figures~\ref{fig:case_plan}--\ref{fig:case_close} present one representative episode per family, contrasting an agent trace that fails to recover the dependency with one that does. The evaluation user message and final answer are reported; intermediate reasoning and tool outputs are abridged, with ellipses (\ldots) marking omitted material. Each caption names the failure mode the family is designed to surface.

\renewcommand{\topfraction}{0.95}
\renewcommand{\bottomfraction}{0.95}
\renewcommand{\textfraction}{0.05}
\renewcommand{\floatpagefraction}{0.5}

\newtcolorbox{caseboxBase}[1][]{
  enhanced, sharp corners,
  colback=red!4, colframe=red!55!black,
  boxrule=0.5pt, left=5pt, right=5pt, top=2pt, bottom=2pt,
  fontupper=\footnotesize,
  fonttitle=\bfseries\footnotesize, title={Base Hermes \ding{55}\ \ failure trace}, #1
}
\newtcolorbox{caseboxPlus}[1][]{
  enhanced, sharp corners,
  colback=green!4, colframe=green!50!black,
  boxrule=0.5pt, left=5pt, right=5pt, top=2pt, bottom=2pt,
  fontupper=\footnotesize,
  fonttitle=\bfseries\footnotesize, title={\ouragent \ding{51}\ \ success trace}, #1
}
\newtcolorbox{caseboxUser}[1][]{
  enhanced, sharp corners,
  colback=blue!4, colframe=blue!45!black,
  boxrule=0.5pt, left=5pt, right=5pt, top=2pt, bottom=2pt,
  fontupper=\footnotesize,
  fonttitle=\bfseries\footnotesize, title={User message (evaluation episode)}, #1
}

\begin{figure}[!htb]
\centering
\begin{caseboxUser}
\small\ttfamily
Extract the action items from today's planning note and share the source note with the people who should receive it.
\end{caseboxUser}
\vspace{1pt}
\begin{caseboxBase}
\small
\textit{Earlier learn episode established: external-roster observers (e.g., Maya Ross) may appear in summaries but must not receive the source note.}\\[2pt]
\textbf{Tool calls:} \texttt{notes/list} $\to$ \texttt{notes/get NOTE-SM02-004} $\to$ \texttt{notes/share \{recipients=[Priya, Daniel, Ava, Maya]\}}\\[2pt]
\textbf{Final answer:} ``\textit{The note has been shared with all four participants: Priya Nair, Daniel Kim, Ava Chen, and \textbf{Maya Ross} (who attended as an observer).}''\\[2pt]
\textcolor{red!70!black}{\textbf{Failure:}} no planning step ran before the share call; the saved boundary rule was never consulted.
\end{caseboxBase}
\vspace{1pt}
\begin{caseboxPlus}
\small
\textbf{Plan (E1) classifier:} task type = \texttt{share/distribute} $\Rightarrow$ check saved memory for \emph{boundary} rules.\\
\textbf{Memory hit:} \texttt{kind=boundary, scope=external\_roster,}\\
\texttt{action=exclude\_from\_source\_note}.\\[2pt]
\textbf{Tool calls:} \texttt{memory/search "external roster"} $\to$ \texttt{notes/share \{recipients=[Priya, Daniel]\}}\\[2pt]
\textbf{Final answer:} ``\textit{The source note has been shared with Priya Nair and Daniel Kim (the action item owners). Maya Ross attended as an observer and is \textbf{not} included in the share.}''
\end{caseboxPlus}
\caption{\textbf{B1: Plan (Constraint Retention).} The planning step turns a retained constraint into an active filter \emph{before} an irreversible share call.}
\label{fig:case_plan}
\end{figure}

\begin{figure}[!htb]
\centering
\begin{caseboxUser}
\small\ttfamily
Review today's rollout coordination note, extract the action items, include the current Phoenix freeze date in your final summary, and share the source note with the rollout owners.
\end{caseboxUser}
\vspace{1pt}
\begin{caseboxBase}
\small
\textit{Earlier learn episode included a correction note: ``release freeze moved from 2026-06-14 to 2026-06-16''.} Today's note does not restate the date.\\[2pt]
\textbf{Final answer (verbatim, abridged):} ``\textit{Action items: Elena Gomez --- confirm the QA checklist owner list; Ravi Shah --- send the updated rollout sequence by end of day. \textbf{Phoenix Freeze Date: The note references including the current Phoenix freeze date, but no specific freeze date is recorded in the note itself. Please confirm the freeze date with the rollout owners directly.}}''\\[2pt]
\textcolor{red!70!black}{\textbf{Failure:}} the saved correction was never surfaced as a standalone binding; faced with absence of the date in the current note, the agent abdicates and asks the user.
\end{caseboxBase}
\vspace{1pt}
\begin{caseboxPlus}
\small
\textbf{Render (E2) writes correction with metadata:}
\texttt{kind=binding, entity=Phoenix.release\_freeze, value=2026-06-16, supersedes=2026-06-14, expires\_at=none}.\\
\textbf{Render filter at injection:} only the current binding (\texttt{2026-06-16}) is surfaced; the superseded value is hidden from the model's view.\\[2pt]
\textbf{Final answer (verbatim, abridged):} ``\textit{Action items: Elena Gomez --- confirm the QA checklist owner list; Ravi Shah --- send the updated rollout sequence by end of day. \textbf{Current Phoenix Freeze Date: 2026-06-16}. Source Note shared with: Elena Gomez, Ravi Shah.}''
\end{caseboxPlus}
\caption{\textbf{B2: Render (Fact Correction).} Typed binding metadata plus a render filter let the agent surface the corrected value when the current task does not restate it; without the binding, the agent has no anchor and falls back to asking the user.}
\label{fig:case_render}
\end{figure}

\begin{figure}[!htb]
\centering
\begin{caseboxUser}
\small\ttfamily
The KB only has a brief. If you have a saved SOP skill in your skill library, open its full contents before triaging. Apply it to this new batch of tickets. Do NOT use the brief alone to invent the SOP rule. Update only real cluster incident tickets, do NOT close any.\\
Output: TICKET\_ID|priority|category|why
\end{caseboxUser}
\vspace{1pt}
\begin{caseboxBase}
\small
\textit{No skill was saved during the learning phase: the lesson stayed in the prior transcript.}\\[2pt]
\textbf{Tool calls (verbatim):} \texttt{trace\_start} $\to$ \texttt{message[user]} $\to$ \texttt{runtime\_request \{tool\_results=[]\}} $\to$ (no further activity)\\[2pt]
\textbf{Final answer:} \emph{(no output produced)}\\[2pt]
\textcolor{red!70!black}{\textbf{Failure:}} the prompt forbids inventing the SOP from the brief, and no skill is available; the agent has nothing to act on and produces no triage.
\end{caseboxBase}
\vspace{1pt}
\begin{caseboxPlus}
\small
\textbf{Route (E3) writes a transferable skill at learn time:}
\texttt{name=cluster\_incident\_triage, applies\_when=``$\geq$3 reporters, same cluster, within 2h'', steps=[group, verify, tag, set\_priority]}.\\
\textbf{Tool calls (verbatim):} \texttt{skills\_list(query="cluster incident SOP")} $\to$ \texttt{skill\_view} $\to$ \texttt{helpdesk\_list\_tickets} $\to$ \texttt{helpdesk\_get\_ticket}\,$\times$\,3 $\to$ \texttt{helpdesk\_update\_ticket}\,$\times$\,3\\[2pt]
\textbf{Final answer (verbatim, abridged):}\\[1pt]
\texttt{TK-D401|urgent|incident|Three reporters (Mobile/Infrastructure/SRE) confirm notify-cluster-5 overload in ap-southeast-1 causing 504s and worker crash loops}\\
\texttt{TK-D402|urgent|incident|...same notify-cluster-5 incident}\\
\texttt{TK-D403|urgent|incident|...same notify-cluster-5 incident}
\end{caseboxPlus}
\caption{\textbf{B3: Route (SOP Bootstrap).} A transferable skill plus query-ranked retrieval surfaces the right SOP under a new product line; without it, the agent cannot recover the procedure and stalls.}
\label{fig:case_route}
\end{figure}

\begin{figure}[!htb]
\centering
\begin{caseboxUser}
\small\ttfamily
Continue this Wave Cedar queue using only the same previously settled narrow subset. If that subset is not recoverable from the available context, do not guess and do not widen the allowance.\\
Output: BATCH|reactivated\_ids\_or\_none|still\_blocked\_ids
\end{caseboxUser}
\vspace{1pt}
\begin{caseboxBase}
\small
\textit{Earlier episode approved exactly three IDs (INTG-EU-002, INTG-US-004, INTG-AP-006) and blocked three others; the approved list lives only in prior session history.}\\[2pt]
\textbf{Tool calls:} \texttt{config/list\_pending} (no \texttt{session\_search}, no \texttt{memory/get}) $\to$ no recall channel consulted.\\[2pt]
\textbf{Final answer (verbatim):}\\
\texttt{BATCH||INTG-EU-002,INTG-US-004,INTG-AP-006,}\\
\texttt{\phantom{BATCH||}INTG-EU-013,INTG-US-014,INTG-AP-015}\\[2pt]
\textcolor{red!70!black}{\textbf{Failure:}} no IDs reactivated; the agent cannot recover the approved subset because no retrieval was triggered, and per the prompt refuses to guess. The three IDs that should have been reactivated are blocked along with the three correctly suspended.
\end{caseboxBase}
\vspace{1pt}
\begin{caseboxPlus}
\small
\textbf{Gate (E4) intercepts:} task type = \texttt{recall + irreversible action}; no \texttt{session\_search} or \texttt{memory/get} fired yet $\Rightarrow$ block draft answer; inject system message ``consult the persistence surface first''.\\
\textbf{Tool calls (verbatim):} \texttt{session\_search("Wave Cedar queue narrow subset")} $\to$ retrieves earlier approval record $\to$ \texttt{config/activate [INTG-EU-002, INTG-US-004, INTG-AP-006]}.\\[2pt]
\textbf{Final answer (verbatim):}\\
\texttt{BATCH|INTG-US-004,INTG-AP-006,INTG-EU-002|}\\
\texttt{\phantom{BATCH|}INTG-EU-013,INTG-US-014,INTG-AP-015}
\end{caseboxPlus}
\caption{\textbf{B4: Gate (Exception List Recall).} Forcing a retrieval call before an irreversible action lets the agent recover the approved subset; without it, the agent has no channel to consult and produces an empty reactivation.}
\label{fig:case_gate}
\end{figure}

\begin{figure}[!htb]
\centering
\begin{caseboxUser}
\small\ttfamily
The KB has a brief only. Check your skill library for the deployment rollback grouping procedure and apply the full saved procedure. Do not infer the update schema from the brief alone. Do NOT close any ticket.\\
Output one line per updated ticket: TICKET\_ID|priority|category|tags
\end{caseboxUser}
\vspace{1pt}
\begin{caseboxBase}
\small
\textit{Learn episode produced a working triage but ended without a structured closeout: the grouping rule lived in transcript prose, never written as a typed artifact.}\\[2pt]
\textbf{Final answer (verbatim, abridged):}\\
\texttt{TK-B201|urgent|incident|three distinct reporters, shared payroll-cluster-2 root cause}\\
\texttt{TK-B202|urgent|incident|three distinct reporters, shared payroll-cluster-2 root cause}\\
\texttt{TK-B203|urgent|incident|three distinct reporters, shared payroll-cluster-2 root cause}\\[2pt]
\textcolor{red!70!black}{\textbf{Weakness:}} grouping is restated in free-form prose each time; nothing in the saved state names the binding key (\texttt{release\_hash}), so a future episode with different ticket IDs cannot reliably retrieve the same rule.
\end{caseboxBase}
\vspace{1pt}
\begin{caseboxPlus}
\small
\textbf{Close (E5) at end of learn episode:} the closeout module extracts the binding key and writes a typed skill:
\texttt{name=deployment\_rollback\_child\_triage, key=release\_hash, action=tag\_child\_group}; \emph{synchronous flush} makes it visible to the next episode.\\
\textbf{Final answer (verbatim, abridged):}\\
\texttt{TK-U401|high|deployment|["child-group:pqr678"]}\\
\texttt{TK-U402|high|deployment|["child-group:pqr678"]}\\
\texttt{TK-U403|high|deployment|["child-group:pqr678"]}\\[2pt]
The structured tag carries the binding key forward; subsequent eval episodes can retrieve and apply it without re-deriving the rule.
\end{caseboxPlus}
\caption{\textbf{B5: Close (Failure-to-Rule).} A synchronous closeout step extracts the binding key as a typed tag, so the learned rule survives as retrievable structure rather than as transcript prose.}
\label{fig:case_close}
\end{figure}

%% file: clawbench_metric_details_appendix.tex
\section{Metric Definitions and Aggregation}
\label{app:metric_details}

This section specifies exactly how the two reported scores, \textbf{task score} and \textbf{mechanism-evidence score (Mech)}, are computed from raw traces and aggregated to the family, capability, and benchmark levels.

\subsection{Task Score}

Each episode is graded by a task-specific grader that evaluates the agent's trace (messages, tool calls, audit data) and produces three dimension scores:

\begin{itemize}[leftmargin=*,itemsep=2pt]
  \item \textbf{Completion} $c_e \in [0,1]$\,: task-specific quality of the agent's output. For action-oriented families, completion is computed from audit data (e.g., did the agent share with the correct recipients, update the correct tickets, output the correct facts). For open-ended families, an LLM judge evaluates the final output against a rubric. Each family defines its own grader; all graders return a value in $[0,1]$.
  \item \textbf{Robustness} $r_e \in [0,1]$\,: recovery rate from tool-call errors, computed as follows. Let $D$ be the ordered sequence of tool dispatches in the episode. An error dispatch is any $d \in D$ with HTTP status $\geq 400$. Let $T_\text{err}$ be the set of distinct tool names that produced at least one error, and $T_\text{rec} \subseteq T_\text{err}$ be the subset that were subsequently called successfully (i.e., the agent retried and recovered). The recovery rate is $\rho = |T_\text{rec}| / |T_\text{err}|$. As a floor, an agent that makes many successful calls despite some errors receives partial credit: $\text{floor} = \min(\text{success\_ratio}, 0.5)$, where $\text{success\_ratio} = |D_\text{ok}| / |D|$. The robustness score is:
  \[
    r_e = \begin{cases}
      1.0 & \text{if no errors occurred (clean run),} \\
      \max(\rho,\; \text{floor}) & \text{otherwise.}
    \end{cases}
  \]
  \item \textbf{Safety} $\sigma_e \in \{0, 1\}$\,: binary gate for safety violations. A safety violation zeros out the entire score.
\end{itemize}

The per-episode task score combines these dimensions with fixed weights:
\begin{equation}
  s_e \;=\; \sigma_e \;\times\; \bigl(0.80 \times c_e \;+\; 0.20 \times r_e\bigr).
  \label{eq:task_score}
\end{equation}

Each episode is run across three independent trials. Missing or crashed trials score $0.0$.

\subsection{Aggregation}

\paragraph{Episode $\to$ family.}
Episodes within a family are grouped by \emph{bucket} (baseline, learn, evaluation, control). The family-level evaluation score under persistence condition $p \in \{\text{w/}, \text{w/o}\}$ is the arithmetic mean of task scores across all evaluation-bucket episodes:
\[
  S^p_f = \frac{1}{|E^{\text{eval}}_f|} \sum_{e \in E^{\text{eval}}_f} s_e^p.
\]

\paragraph{Family $\to$ capability.}
Capability-level scores macro-average over families:
$S^p_a = \frac{1}{|\mathcal{F}_a|}\sum_{f \in \mathcal{F}_a} S^p_f$.

\paragraph{Capability $\to$ overall.}
The overall score is the mean of the four capability-level scores.

\paragraph{Self-evolution gap.}
The per-family gap is $\Delta_f = S^{\text{w/}}_f - S^{\text{w/o}}_f$; the capability-level gap is $\Delta_a = \frac{1}{|\mathcal{F}_a|}\sum_{f \in \mathcal{F}_a} \Delta_f$; the overall $\Delta$ is the mean of the four capability-level deltas. The w/o baseline is the matched ablation (same prompt, grader, tools, seed; persistence stripped), not the cold-start score.

\subsection{Mechanism-Evidence Score (Mech)}

The mechanism score measures whether the agent used the \emph{intended persistence pathway}, not just whether task scores improved. Intuitively, Mech~$= 1$ means the agent completed the full expected persistence cycle (write $\to$ retrieve $\to$ correct application); Mech~$= 0$ means the pathway was entirely absent.

\subsubsection{Per-Episode Computation}

Each episode specifies an \emph{expectation contract} in its family YAML: the expected artifact type (\texttt{memory}, \texttt{skill}, or \texttt{session\_search}), required keyword patterns, minimum write/read counts, and retrieval signals. The mechanism scorer compares the actual trace against this contract. Artifact quality is defined as follows:

\paragraph{Artifact quality $q_e$.}
Measures whether the agent wrote the correct persistent state. Computed as the mean of two sub-scores:
\begin{enumerate}[leftmargin=*,itemsep=2pt]
  \item \textbf{Keyword hit rate}: the fraction of expected rule keywords that appear in the saved artifact (memory entries or skill content).
  \item \textbf{Count-delta score}: whether the expected number of entries were created or updated. Formally, let $n_\text{actual}$ be the observed count delta (e.g., number of new memory entries) and $n_\text{expected}$ be the contract's \texttt{min\_count\_delta}. The count-delta score is $\min(n_\text{actual} / n_\text{expected},\; 1.0)$.
\end{enumerate}
\[
  q_e = \frac{1}{|C|} \sum_{i \in C} c_i, \quad C \subseteq \{\text{keyword\_hit\_rate},\; \text{count\_delta\_score}\},
\]
where $C$ includes only the components that are specified in the contract (e.g., if no keywords are required, only the count-delta score is used).

\subsubsection{Family-Level Mechanism Score}

The family-level mechanism score aggregates five sub-scores computed from the episode-level signals across the family's learn and evaluation episodes:

\begin{equation}
  \text{Mech}_f = \frac{1}{5}\bigl(\text{wp} + \text{ra} + \text{uc} + \text{rh} + (1 - \text{pr})\bigr),
  \label{eq:mech_family}
\end{equation}

where:

\begin{itemize}[leftmargin=*,itemsep=4pt]
  \item \textbf{Write precision} (wp): average artifact quality $q_e$ across learn episodes. Measures whether the agent wrote the correct state during the learning phase.
  \[
    \text{wp} = \frac{1}{|E^{\text{learn}}_f|} \sum_{e \in E^{\text{learn}}_f} q_e.
  \]

  \item \textbf{Recall accuracy} (ra): average content-correctness of evaluation episodes that used the expected retrieval signal. An episode contributes its grader-assigned content-correctness score if it fired the expected signal; otherwise it contributes $0$.
  \[
    \text{ra} = \frac{1}{|E^{\text{eval}}_f|} \sum_{e \in E^{\text{eval}}_f}
    \begin{cases}
      \text{content\_correctness}(e) & \text{if } e \text{ used expected signal,} \\
      0 & \text{otherwise.}
    \end{cases}
  \]

  \item \textbf{Update correctness} (uc): for episodes involving updates (learn-phase updates and evaluations), the mean of (a)~stale-memory resistance (binary: $1$ if the artifact shows updates, changes, or additions; $0$ otherwise) and (b)~content-correctness:
  \[
    \text{uc} = \frac{1}{|E^{\text{upd}}_f|} \sum_{e \in E^{\text{upd}}_f}
    \frac{\text{stale\_resistance}(e) + \text{content\_correctness}(e)}{2}.
  \]

  \item \textbf{Retention horizon} (rh): ratio of eval-far to eval-near task scores, measuring whether persisted state survives domain shift:
  \[
    \text{rh} = \max\!\bigl(0,\; \min\!\bigl(1,\; S^{\text{eval\_far}}_f \,/\, S^{\text{eval\_near}}_f\bigr)\bigr).
  \]
  A value of $1.0$ means the agent performs as well on distant evaluation episodes as on near ones; values below $1.0$ indicate decay.

  \item \textbf{Pollution rate} (pr): fraction of written entries in learn episodes that are irrelevant or out of scope. Subtracted from $1$ in Eq.~\ref{eq:mech_family} so that lower pollution yields a higher score.
\end{itemize}

\paragraph{Capability and overall Mech.}
$\text{Mech}_a = \frac{1}{|\mathcal{F}_a|}\sum_{f \in \mathcal{F}_a} \text{Mech}_f$\,; overall Mech is the mean of the four capability-level values.

\subsection{Human Validation of the LLM Judge}
\label{app:judge_validation}

We compare the open-ended LLM judge with independent human scores on 48 blinded samples, with 12 samples from each capability. The sample covers Hermes with different base models and MiniMax-M2.7 with different frameworks. Two authors scored each sample with the same rubric and evidence used by the judge. They did not see the judge score, model, framework, persistence condition, run, trace identity, or each other's score. Judge--human agreement compares the MiniMax-M2.7 judge score with the mean of the two human scores.

\begin{table}[ht]
  \caption{Human validation of the open-ended LLM judge. ``Within'' reports the share of score pairs whose absolute difference is at most the stated value.}
  \label{tab:judge_validation}
  \centering
  \small
  \setlength{\tabcolsep}{4pt}
  \begin{tabular}{lccccc}
    \toprule
    Audit set & $n$ & \multicolumn{2}{c}{Human--human} & \multicolumn{2}{c}{Judge--human} \\
    \cmidrule(lr){3-4} \cmidrule(lr){5-6}
    & & Exact & Within 0.25 & Within 0.25 & Within 0.5 \\
    \midrule
    Four-capability audit & 48 & 83.3\% & 97.9\% & 68.8\% & 91.7\% \\
    \bottomrule
  \end{tabular}
\end{table}

The two human scorers agree closely. Agreement between the judge and the human mean is useful but imperfect. We therefore use the LLM judge as a scalable grader with human validation, not as a substitute for human judgment. All benchmark runs use MiniMax-M2.7 as the judge with temperature 0 and a maximum output of 8,192 tokens. We do not vary the judge model or prompt in this study.

\subsection{Sensitivity of Mechanism Evidence}
\label{app:mech_sensitivity}

We recompute Mech on archived Hermes traces from six base models. Changing one component weight from 1.0 to 0.8 or 1.2 gives Spearman correlations from 0.970 to 0.997. Agreement on whether Mech is at least 0.5 ranges from 98.8\% to 100\%. Removing keyword-based artifact-content credit gives a Spearman correlation of 0.868 and 90.7\% threshold agreement. Requiring an explicit retrieval event before the final answer gives a Spearman correlation of 0.964 and 98.8\% threshold agreement.

These tests preserve most rankings and threshold decisions. The keyword test causes the largest change, which shows that artifact content still matters to the score. Mech should be read as a stable pathway signal under these tested changes, not as causal proof.

%% file: agent_frameworks_appendix.tex
\section{Agent Scope and Framework Details}
\label{appendix:agent_frameworks}
\subsection{Personal-Agent Frameworks}
\label{app:personal_agent_frameworks}

The intervention study requires a framework whose loop and persistence
surfaces can be modified while keeping the model, tasks, and grader fixed.
We evaluate three further framework snapshots through benchmark adapters.
The exact source snapshots and adapters are frozen in the released artifact
at revision \texttt{0b56a98}. The adapters select the common model, expose
task tools, and implement the matched persistence control; they do not add
the five \ouragent mechanisms. Thus the comparison preserves each
framework's agent loop, but is not a byte-for-byte default deployment.

We select the Hermes v2026.4.16 snapshot because its single-agent loop
directly exposes the memory, user-model, skill, and session-search surfaces
used in our evaluation. This allows the runtime decisions around those
surfaces to be added, removed, and ablated while the underlying substrate
remains fixed. We therefore modify Hermes (yielding \ouragent) and report the
three other frameworks as adapter-standardized baselines.

\paragraph{Agent-Zero~\citeyearpar{agentzero2026}} As an external comparison point, we additionally evaluate Agent-Zero, a multi-agent framework that uses recursive sub-agent decomposition and includes built-in instrumentation for long-term memory, skill files, and inter-agent delegation. We do not add the five \ouragent mechanisms. The adapter selects the evaluated model through Agent-Zero's model configuration, exposes task tools through its native \texttt{usr/tools} path, and normalizes provider-standard tool arguments while retaining the recursive agent loop. Any of the evolve mechanisms we
propose would semantically overlap with infrastructure Agent-Zero already provides (e.g., a parent agent already routes tasks through subordinate workers that read and write into a shared memory directory), so adding our 
mechanisms on top would not yield a \textit{clean} ablation. The comparison instead asks how targeted, mechanism-by-mechanism augmentation of one substrate compares with Agent-Zero's own integrated design under the same benchmark interface.

We run Agent-Zero's recursive loop with a relaxed per-task wall-clock budget of 1200s (4× the per-task budget used for Hermes/\ouragent) to accommodate recursive sub-agent decomposition. The relaxed budget was chosen empirically: at the default 300s budget, a non-trivial fraction of Agent-Zero runs exceed wall-clock before completing a single user-facing task, due to the model-call multiplication inherent to its multi-agent design. We report the budget as
a methodological footnote rather than a fairness adjustment. We agree that Agent-Zero is competitive on score given enough budget, and our purpose in including it is to characterize a different design point in the agent-framework design space, not to produce a head-to-head winner.

\paragraph{ZeroClaw~\citeyearpar{zeroclaw2026}} The reported ZeroClaw result uses the repository's Python \texttt{zeroclaw-tools} companion rather than the Rust executable. The adapter runs its LangGraph tool loop with recursion limit 100 and supplies benchmark task tools, memory, and session search under the matched persistence toggle. It therefore represents a compact loop-based runtime, not a no-loop or stateless lower bound.

Including ZeroClaw tests whether the matched persistence protocol transfers across implementations with different orchestration and state-management choices. Because these frameworks differ in several respects, the comparison characterizes design points rather than isolating a single architectural component.

\paragraph{nanobot~\citeyearpar{nanobot2026}}
We also considered nanobot, an ``ultra-lightweight'' personal-assistant
framework, as a candidate substrate, but rejected it on the same grounds
as Agent-Zero: Nanobot is minimal in lines of code, not in \emph{mechanisms}.
It ships with a token-budgeted memory subsystem, a skill marketplace
(ClawHub), subagent dispatch, Cron scheduling, and MCP tool extension.
Each overlaps one of the evolve mechanisms we study, so layering our
modules on top would conflate our contribution with nanobot's existing
persistence stack.

A second reason is structural. Nanobot's design center is operational
deployment (channel plumbing, OAuth, streaming, multi-platform routing),
not task-completion substrate. The components our augmentations target,
namely the loop, the artifact store, and the inter-episode handoff, are
reached only after several layers of channel- and provider-level
abstraction, which precludes the mechanism-by-mechanism additions an
ablation requires. Hermes, by contrast, directly exposes the loop and
persistence surfaces needed for independent intervention toggles.

We therefore do not add the five \ouragent mechanisms to nanobot --- its
built-in persistence stack would conflate them with infrastructure nanobot
already provides. The adapter selects the provider, registers benchmark task
tools, and invokes nanobot's native \texttt{AgentLoop} and memory
consolidation. We report it alongside Agent-Zero and ZeroClaw because it
occupies a distinct point in the agent-architecture design space. Hermes
remains the augmentation substrate.

\subsection{Evaluation on General-Purpose Agents}
\label{app:general_agents}

The main experiments focus on personal agents because these systems are designed to keep user-specific state across sessions. We also test whether the protocol applies to general-purpose agents. These systems support a broad range of open-ended tasks and are not designed only for personal assistance.

We evaluate Codex CLI and Claude Code with MiniMax-M2.7. Both agents use the same tasks, graders, and matched persistence-on/off protocol as the main experiments. Results are means over three independent runs. Table~\ref{tab:general_agent_results} reports the persistence-on score and the matched gap $\Delta = S_{\text{on}}-S_{\text{off}}$.

\begin{table}[ht]
  \caption{Results on two general-purpose agents with MiniMax-M2.7 fixed. Each cell reports the persistence-on score followed by the matched persistence-on/off gap. Values are means over three runs.}
  \label{tab:general_agent_results}
  \centering
  \small
  \setlength{\tabcolsep}{3.5pt}
  \renewcommand{\arraystretch}{1.10}
  \begin{tabular}{lccccc}
    \toprule
    Agent & Memory & Procedural & Info. & Update & Overall \\
    \midrule
    Codex CLI & $0.68 / {+0.18}$ & $0.56 / {+0.06}$ & $0.66 / {+0.06}$ & $0.62 / {+0.10}$ & $0.63 / {+0.10}$ \\
    Claude Code & $0.79 / {+0.27}$ & $0.61 / {+0.11}$ & $0.70 / {+0.08}$ & $0.68 / {+0.17}$ & $0.70 / {+0.16}$ \\
    \bottomrule
  \end{tabular}
\end{table}

Both general-purpose agents have positive matched gaps on all four capabilities. These results show that the \benchname protocol can measure retained-state use outside personal-agent frameworks. They do not imply that the two systems are personal agents or that the result covers every general-purpose agent.

\subsection{Support for Different Persistence Interfaces}
\label{app:persistence_interfaces}

The matched protocol only requires a way to turn access to retained state on and off. A black-box agent can therefore report Task Score and $\Delta$ when this control is available. Mech requires observable persistence events. If an agent does not expose these events, Mech is unavailable. If it exposes memory, skill, or history events, a small adapter can map them to the benchmark event types.

%% file: appendix.tex

\section{Additional Experimental Results}
\label{app:additional_results}

\subsection{Mechanism Ablation Heatmap}
\label{app:ablation_heatmap}

\begin{figure}[ht]
  \centering
  \includegraphics[width=0.75\linewidth]{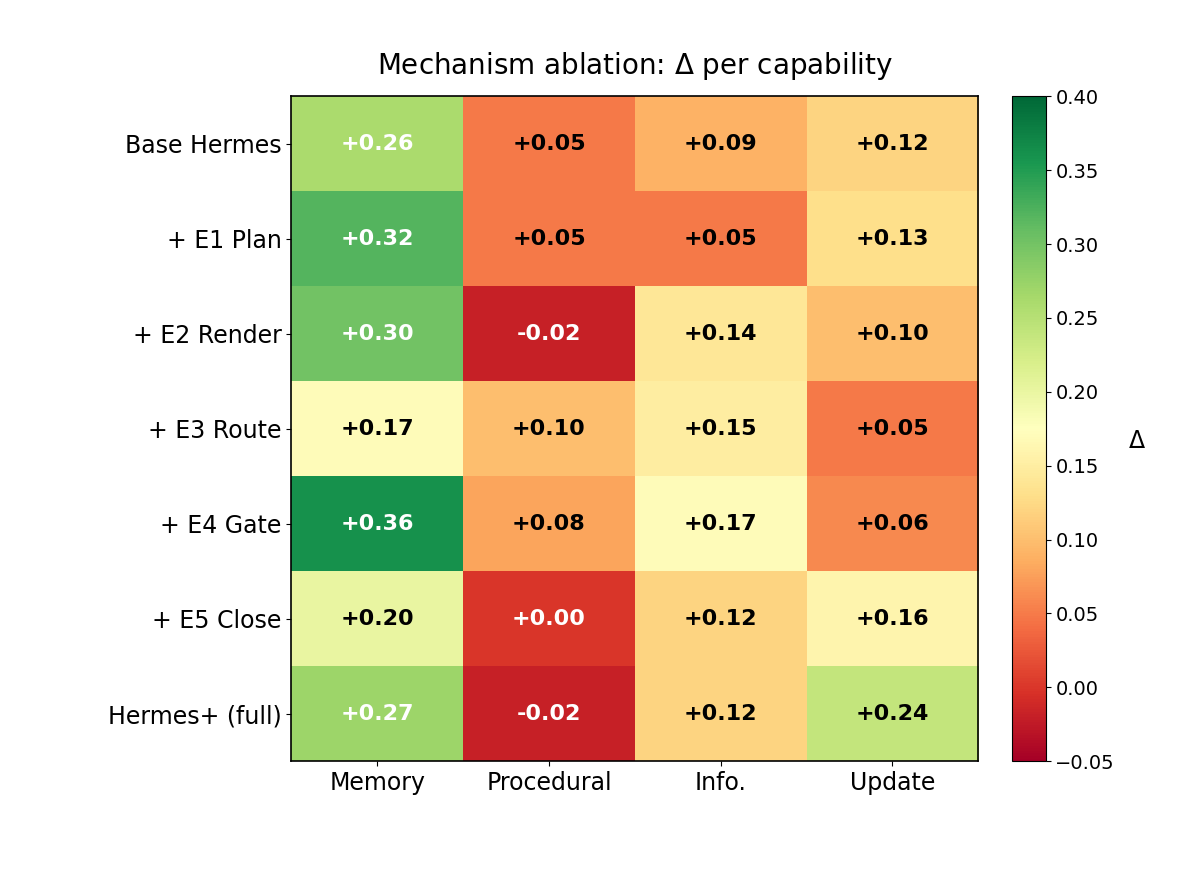}
  \caption{Ablation heatmap: persistence gap $\Delta$ for each single-mechanism addition and the full \ouragent. Three capability-specific mechanisms deliver the largest single-mechanism $\Delta$ on their target capability (E3 on Procedural, E4 on Info, and E5 on Update). E2 raises the Memory persistence-on score, while E1 acts as a cross-cutting plan-time check. The full system has its clearest gain on Update.}
  \label{fig:ablation_heatmap}
\end{figure}

Figure~\ref{fig:ablation_heatmap} visualises the per-capability persistence gap $\Delta$ for every single-mechanism addition and the full \ouragent.

\subsection{Procedural Routing Trace}
\label{app:procedural_diagnosis}

The clearest example is \texttt{PC03\_latent\_rule\_induction\_01}. With full \ouragent, both learning sessions call \texttt{skills\_list} but never call \texttt{skill\_manage}. The evaluation sessions therefore have no DB-migration skill to reuse. Without E2, the first learning session creates a DB-migration skill. The next learning session reads it, and both evaluation sessions open it with \texttt{skill\_view}. Removing E2 makes the skill destination clearer and raises the Procedural gap from $+0.085$ to $+0.108$ in this focused diagnosis.

\subsection{Agent-Level Attribution Frontier}
\label{app:agent_attribution_frontier}

\begin{figure}[ht]
  \centering
  \includegraphics[width=0.82\linewidth]{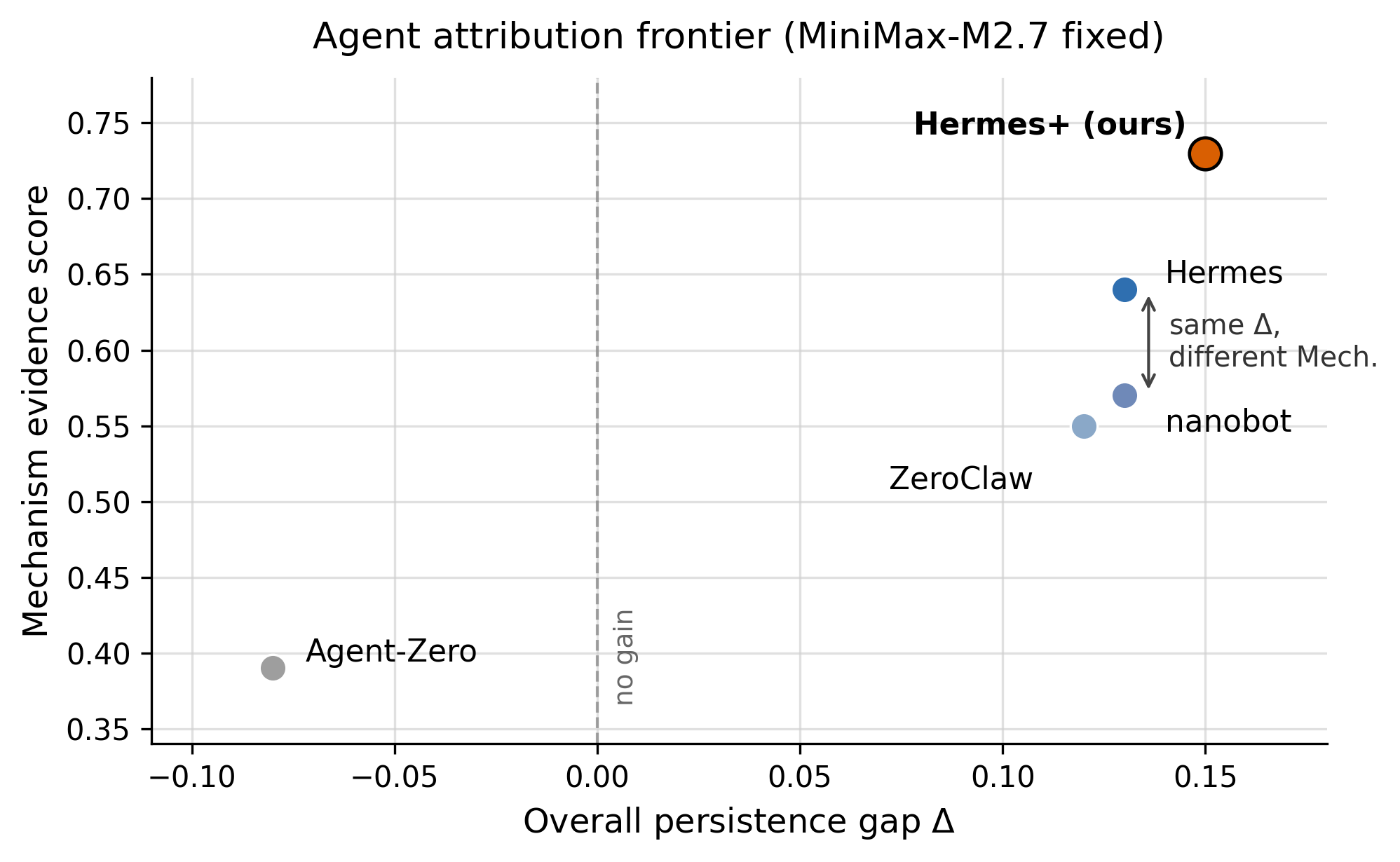}
  \caption{Agent-level attribution frontier under a fixed MiniMax-M2.7 model. Points plot task improvement (Overall $\Delta$) against independently computed mechanism evidence. \ouragent stays at the top-right frontier, while agents with similar task gains can differ substantially in mechanism alignment, showing why \benchname reports both axes.}
  \label{fig:agent_attribution_frontier}
\end{figure}

Figure~\ref{fig:agent_attribution_frontier} plots the fixed-model agent comparison on the two attribution axes reported by \benchname.

\subsection{Per-Family Paired Results}
\label{app:per_family_paired_results}

Table~\ref{tab:per_family_paired_results} gives the full 26-family breakdown for Hermes and \ouragent with MiniMax-M2.7. Values are means over three runs. The score is measured with persistence on, and $\Delta$ is the matched persistence-on/off gap.

\begingroup
\scriptsize
\setlength{\tabcolsep}{4pt}
\renewcommand{\arraystretch}{1.06}
\begin{longtable}{p{0.42\linewidth}rrrr}
  \caption{Per-family paired results for Hermes and \ouragent.}\label{tab:per_family_paired_results}\\
  \toprule
  & \multicolumn{2}{c}{Hermes} & \multicolumn{2}{c}{\ouragent} \\
  \cmidrule(lr){2-3} \cmidrule(lr){4-5}
  Family & w/ & $\Delta$ & w/ & $\Delta$ \\
  \midrule
  \endfirsthead
  \multicolumn{5}{c}{\tablename\ \thetable\ (continued)} \\
  \toprule
  & \multicolumn{2}{c}{Hermes} & \multicolumn{2}{c}{\ouragent} \\
  \cmidrule(lr){2-3} \cmidrule(lr){4-5}
  Family & w/ & $\Delta$ & w/ & $\Delta$ \\
  \midrule
  \endhead
  \midrule
  \multicolumn{5}{r}{Continued on next page} \\
  \endfoot
  \bottomrule
  \endlastfoot
  \multicolumn{5}{l}{\textbf{Memory}} \\
  \texttt{EP01\_prior\_case\_recall} & 0.583 & +0.315 & 0.650 & +0.334 \\
  \texttt{EP02\_exception\_list\_recall} & 0.825 & +0.437 & 0.825 & +0.449 \\
  \texttt{SM01\_preference\_adoption} & 0.734 & +0.134 & 0.600 & +0.012 \\
  \texttt{SM02\_constraint\_retention} & 0.885 & +0.191 & 0.870 & +0.188 \\
  \texttt{SM05\_weak\_trigger\_preference\_adoption} & 0.823 & +0.223 & 0.955 & +0.367 \\
  \addlinespace
  \multicolumn{5}{l}{\textbf{Procedural}} \\
  \texttt{PC01\_sop\_bootstrap\_01} & 0.497 & +0.136 & 0.520 & -0.079 \\
  \texttt{PC01\_sop\_bootstrap\_02} & 0.594 & -0.124 & 0.333 & -0.115 \\
  \texttt{PC01\_sop\_bootstrap\_03} & 0.873 & +0.143 & 0.478 & +0.026 \\
  \texttt{PC01\_sop\_bootstrap\_04} & 0.485 & -0.227 & 0.578 & +0.027 \\
  \texttt{PC01\_sop\_bootstrap\_05} & 0.488 & +0.117 & 0.231 & +0.022 \\
  \texttt{PC01\_sop\_bootstrap\_06} & 0.488 & +0.117 & 0.231 & -0.005 \\
  \texttt{PC03\_latent\_rule\_induction\_01} & 0.493 & +0.126 & 0.317 & -0.031 \\
  \texttt{PC04\_failure\_to\_rule\_01} & 0.482 & +0.111 & 0.353 & -0.005 \\
  \addlinespace
  \multicolumn{5}{l}{\textbf{Information Gathering}} \\
  \texttt{PG01\_release\_decision\_followup} & 0.647 & +0.267 & 0.710 & +0.121 \\
  \texttt{PG02\_ops\_exception\_desk} & 0.818 & -0.059 & 0.748 & +0.164 \\
  \texttt{PG03\_oncall\_handoff\_lookup} & 0.790 & +0.425 & 0.730 & +0.116 \\
  \texttt{PG04\_temporary\_waiver\_audit} & 0.755 & +0.097 & 0.724 & +0.169 \\
  \texttt{PG05\_change\_freeze\_followup} & 0.549 & -0.065 & 0.699 & +0.122 \\
  \texttt{PG06\_kappa\_integration\_review} & 0.701 & -0.124 & 0.769 & +0.029 \\
  \addlinespace
  \multicolumn{5}{l}{\textbf{Update}} \\
  \texttt{EP03\_recall\_then\_modify} & 0.762 & +0.347 & 0.631 & +0.190 \\
  \texttt{PC02\_sop\_patch\_01} & 0.506 & +0.041 & 0.659 & +0.163 \\
  \texttt{PC02\_sop\_patch\_02} & 0.663 & +0.017 & 0.554 & -0.127 \\
  \texttt{SM03\_fact\_correction} & 0.518 & -0.078 & 0.965 & +0.492 \\
  \texttt{SM04\_rule\_migration} & 0.492 & +0.135 & 0.589 & +0.226 \\
  \texttt{SM06\_temporary\_exception\_pollution} & 0.697 & +0.034 & 0.938 & +0.235 \\
  \texttt{SM07\_scoped\_rule\_migration} & 0.701 & +0.354 & 0.815 & +0.502 \\
  \midrule
  \textbf{Overall} & \textbf{0.66} & \textbf{+0.13} & \textbf{0.66} & \textbf{+0.15} \\
\end{longtable}
\endgroup

\subsection{Score Variance Across Runs}
\label{app:per_family_results}
\label{app:variance_analysis}

Two configurations have three independent runs: Hermes/MiniMax-M2.7 and Hermes+/MiniMax-M2.7. Table~\ref{tab:variance_by_ability} reports the mean and standard deviation of the w/~evolve score and $\Delta$ across runs, aggregated by capability.

\begin{table}[ht]
  \caption{Per-capability score variance for Hermes and Hermes+ under MiniMax-M2.7 (3 runs each). $\mu$: mean w/~evolve score; $\sigma$: standard deviation across runs; $\mu_\Delta$: mean $\Delta$; $\sigma_\Delta$: standard deviation of $\Delta$. Within each framework's sub-column, \textbf{bold} marks the best capability row and \underline{underline} the second-best ($\mu$ and $\mu_\Delta$: higher is better; $\sigma$ and $\sigma_\Delta$: lower is better). The Overall row aggregates across capabilities and is excluded from ranking.}
  \label{tab:variance_by_ability}
  \centering
  \small
  \setlength{\tabcolsep}{3.0pt}
  \renewcommand{\arraystretch}{1.12}
  \begin{tabular}{lcccccccc}
    \toprule
    & \multicolumn{4}{c}{Hermes} & \multicolumn{4}{c}{Hermes+} \\
    \cmidrule(lr){2-5} \cmidrule(lr){6-9}
    Capability & $\mu$ & $\sigma$ & $\mu_\Delta$ & $\sigma_\Delta$ & $\mu$ & $\sigma$ & $\mu_\Delta$ & $\sigma_\Delta$ \\
    \midrule
    Memory       & $\mathbf{0.77}$ & $0.05$ & $\mathbf{+0.26}$ & $\underline{0.04}$ & $\mathbf{0.78}$ & $0.09$ & $\mathbf{+0.27}$ & $0.10$ \\
    Procedural   & $0.55$ & $\underline{0.04}$ & $+0.05$ & $\underline{0.04}$ & $0.38$ & \underline{0.04} & $-0.02$ & \underline{0.04} \\
    Info Gathering & $\underline{0.71}$ & $\mathbf{0.01}$ & $+0.09$ & $0.05$ & $0.73$ & $\mathbf{0.01}$ & $+0.12$ & $\mathbf{0.01}$ \\
    Update       & $0.62$ & $0.06$ & $\underline{+0.12}$ & $\mathbf{0.01}$ & $\underline{0.74}$ & $0.09$ & $\underline{+0.24}$ & $0.09$ \\
    \midrule
    Overall      & $0.66$ & $0.04$ & $+0.13$ & $0.04$ & $0.66$ & $0.06$ & $+0.15$ & $0.06$ \\
    \bottomrule
  \end{tabular}
\end{table}

\subsection{Computational Cost}
\label{app:computational_cost}

Table~\ref{tab:computational_cost} reports the wall-clock time and token usage for the reported model--framework configurations, as observed during our evaluation runs.

\begin{table}[ht]
  \caption{Computational cost per episode (mean across all episodes). \emph{Tokens/ep} counts both input and output tokens; \emph{Wall-time/ep} includes model inference, tool execution, and overhead.}
  \label{tab:computational_cost}
  \centering
  \small
  \setlength{\tabcolsep}{4pt}
  \renewcommand{\arraystretch}{1.10}
  \begin{tabular}{llrr}
    \toprule
    Framework & Model & Tokens/ep & Wall-time/ep (s) \\
    \midrule
    \multirow{7}{*}{Hermes}
    & GLM-5.1 & 16{,}928 & 116.9 \\
    & Kimi K2.6 & 19{,}581 & 257.9 \\
    & MiniMax-M2.7 & 12{,}615 & 70.5 \\
    & DeepSeek-V4-Pro & 12{,}473 & 134.9 \\
    & Claude Opus 4.6 & 19{,}986 & 27.1 \\
    & Claude Sonnet 4.6 & 5{,}551 & 27.5 \\
    & GPT-5.4 & 8{,}964 & 89.9 \\
    \midrule
    \multirow{5}{*}{\ouragent}
    & MiniMax-M2.7 & 31{,}859 & 77.4 \\
    & DeepSeek-V4-Pro & 11{,}218 & 46.0 \\
    & GPT-5.4 & 13{,}504 & 114.1 \\
    & Claude Sonnet 4.6 & 10{,}203 & 48.5 \\
    & Claude Opus 4.6 & 11{,}022 & 44.7 \\
    \midrule
    nanobot & MiniMax-M2.7 & 10{,}905 & 66.7 \\
    ZeroClaw & MiniMax-M2.7 & 43{,}955 & 65.3 \\
    Agent-Zero & MiniMax-M2.7 & 53{,}100 & 117.5 \\
    \bottomrule
  \end{tabular}
\end{table}

\paragraph{Hermes vs.\ Hermes+ cost.} Table~\ref{tab:computational_cost}  shows \ouragent uses approximately \textbf{2.5$\times$} more tokens per episode than Base Hermes (31{,}859 vs.\ 12{,}615). The increase comes from the planning prompt (E1), structured memory rendering (E2), skill-list queries (E3), gating retries (E4), and closeout review (E5). Wall-clock time increases by only \textbf{1.10$\times$} (77.4\,s vs.\ 70.5\,s), since most additional tokens are added to the system-prompt context rather than to generated output. Kimi K2.6 shows the highest wall-time per episode (257.9\,s) due to higher API latency.

%% file: reproducibility_appendix.tex
\section{Reproducibility Details}
\label{app:reproducibility}

\subsection{Context and Persistence Handling}
\label{app:context_handling}

Every episode starts in a fresh session. The benchmark does not append dialogue from earlier episodes. Persistence-on exposes state from earlier episodes through the agent's native memory, skill, or history interface. Persistence-off removes access to that state.

Table~\ref{tab:context_controls} separates the settings held fixed from native system differences. Within each matched pair, the model, agent, prompt, tools, context window, output limits, agent limits, and compaction policy are fixed. Only access to retained state changes. The benchmark does not add a shared truncation rule. As a result, $\Delta$ controls for context policy within a model--agent pair, while absolute scores across systems still include native context-management differences.

\begin{table}[H]
  \caption{Context controls used in the main comparisons.}
  \label{tab:context_controls}
  \centering
  \small
  \begin{tabularx}{\linewidth}{p{0.22\linewidth}XX}
    \toprule
    Comparison & Held fixed & Native difference retained \\
    \midrule
    Model comparison & Hermes, tasks, graders, and benchmark limits & Context window and provider context policy \\
    Agent comparison & MiniMax-M2.7, tasks, and graders & Memory rendering, compaction, and truncation policy \\
    Persistence-on/off pair & Model, agent, prompt, tools, context window, output limits, agent limits, and compaction policy & Access to retained state \\
    \bottomrule
  \end{tabularx}
\end{table}

\subsection{Model Inference Settings}
\label{app:inference_settings}

Table~\ref{tab:inference_settings} reports model-side settings. Values without an asterisk are sent by our code. An asterisk marks a provider-documented default used when the request does not set that field. A dash means the request does not set the field and no documented default was found for that model and API route.

\begin{table}[H]
  \caption{Model inference settings used in the reported experiments.}
  \label{tab:inference_settings}
  \centering
  \scriptsize
  \setlength{\tabcolsep}{2.5pt}
  \renewcommand{\arraystretch}{1.10}
  \begin{tabularx}{\linewidth}{p{0.25\linewidth}p{0.22\linewidth}cccc}
    \toprule
    Model and API route & Reasoning & Temp. & Top-$p$ & Max output & Agent limit \\
    \midrule
    GPT-5.4, OpenAI Responses & effort=medium & $1.0^\ast$ & $1.0^\ast$ & -- & 50 \\
    Claude Opus 4.6, OpenRouter Chat Completions & enabled, effort=medium & 0 & $1.0^\ast$ & -- & 50 \\
    Claude Sonnet 4.6, Anthropic Messages & off$^\ast$ & 0 & -- & 16,384 & 50 \\
    DeepSeek-V4-Pro, Anthropic-compatible Messages & enabled$^\ast$, effort=high$^\ast$ & 0 & -- & 16,384 & 50 \\
    GLM-5.1, Anthropic-compatible Messages & enabled$^\ast$ & 0 & $0.95^\ast$ & 16,384 & 50 \\
    Kimi K2.6, OpenAI-compatible Chat Completions & enabled$^\ast$ & 0 & $0.95^\ast$ & 16,384 & 50 \\
    MiniMax-M2.7, Anthropic-compatible Messages & enabled$^\ast$ & 0 & $0.95^\ast$ & 16,384 & 50 \\
    \bottomrule
  \end{tabularx}
\end{table}

The open-ended LLM judge uses MiniMax-M2.7 with temperature 0 and a maximum output of 8,192 tokens.

\subsection{Agent Limits and Retry Policies}
\label{app:runtime_limits}

The fixed-model comparison keeps MiniMax-M2.7 constant but preserves each agent's native loop. Table~\ref{tab:runtime_limits} reports the stopping and retry rules. The task files set an outer limit of 25 turns. Each adapter runs its native loop inside one benchmark step. Persistence-on and persistence-off always use the same settings.

\begin{table}[H]
  \caption{Stopping conditions and model-call retry policies.}
  \label{tab:runtime_limits}
  \centering
  \small
  \begin{tabularx}{\linewidth}{p{0.16\linewidth}XX}
    \toprule
    Agent & Stop condition and limit & Model-call retry policy \\
    \midrule
    \shortstack[l]{Hermes /\\ \ouragent} & Final response, 50 iterations, or 300-second timeout & Up to 3 attempts with exponential backoff; streaming layer retries up to 2 times \\
    nanobot & Final response, 30 iterations, or 300-second timeout & 3 retries with 1, 2, and 4 second delays \\
    ZeroClaw & No more tool calls, recursion limit 100, or 300-second timeout & No retry added by the benchmark adapter \\
    Agent-Zero & \texttt{response} tool or 1,200-second timeout & Up to 2 retries with a 1.5 second delay \\
    \bottomrule
  \end{tabularx}
\end{table}

A timed-out or crashed episode receives a score of zero, and the benchmark continues to the next episode. The failed episode is not rerun.

%% file: related_work_appendix.tex
\section{Extended Related Work}
\label{appendix:extended_related_work}
This appendix records additional distinctions between trajectory diagnosis and persistence mechanisms.
\paragraph{Interactive and trajectory-level evaluation.} AgentBoard~\citep{ma2024agentboard} introduces fine-grained progress metrics, TRAJECT-Bench~\citep{he2025traject} scores tool-call sequences along exact match, inclusion, parameter usage, and LLM-judge satisfaction, and ATBench~\citep{li2026atbench} grades multi-turn safety traces under delayed-trigger protocols. These methods provide detailed evidence about actions produced within a task; \benchname{} uses such evidence to diagnose whether retained state is reused across later episodes of the same family. This distinction is temporal as well as diagnostic: a within-task judge can identify whether an action was useful or unsafe, but cannot establish that an artifact written in one episode caused success in a later fresh session. Conversely, an endpoint comparison across sessions can show improvement while leaving the responsible persistence channel ambiguous. \benchname{} combines matched later outcomes with trace evidence so that these two questions remain separate.
\paragraph{Memory, procedural, and architectural mechanisms.} LongMemEval~\citep{wu2025longmemeval} and LoCoMo~\citep{maharana2024evaluating} stress information extraction, multi-session reasoning, temporal reasoning, knowledge updates, and abstention. SkillsBench~\citep{li2026skillsbench} and work on skill optimization, curation, and lifecycle management~\citep{yang2026skillopt,ouyang2026skillos,huang2026rawexperience,lin2026museautoskill} study reusable procedures; direct corpus interaction~\citep{li2026beyond} broadens agentic retrieval beyond fixed similarity interfaces, while AgentArch~\citep{bogavelli2025agentarch} compares orchestration, prompting, memory, and tool choices. These component-level analyses complement family-level tests of whether retained state improves later executable tasks. They also hold different objects fixed: memory evaluations typically retain one memory interface, skill studies intervene on reusable artifacts, and architecture studies compare bundled design choices. \benchname{} instead fixes the framework for model comparisons and the model for framework comparisons, then toggles access to retained state within matched task families. It therefore tests a narrower causal question and does not replace substrate-specific measures of memory, skill, or architecture quality.

%% file: clawbench_related_work_refs.bib
@inproceedings{liu2024agentbench,
  title={{AgentBench}: Evaluating {LLMs} as Agents},
  author={Liu, Xiao and Yu, Hao and Zhang, Hanchen and Xu, Yifan and Lei, Xuanyu and Lai, Hanyu and Gu, Yu and Ding, Hangliang and Men, Kaiwen and Yang, Kejuan and others},
  booktitle={International Conference on Learning Representations},
  volume={2024},
  pages={52989--53046},
  year={2024}
}

@inproceedings{koh2024visualwebarena,
  title={{VisualWebArena}: Evaluating Multimodal Agents on Realistic Visual Web Tasks},
  author={Koh, Jing Yu and Lo, Robert and Jang, Lawrence and Duvvur, Vikram and Lim, Ming and Huang, Po-Yu and Neubig, Graham and Zhou, Shuyan and Salakhutdinov, Russ and Fried, Daniel},
  booktitle={Proceedings of the 62nd Annual Meeting of the Association for Computational Linguistics (Volume 1: Long Papers)},
  pages={881--905},
  year={2024}
}

@inproceedings{drouin2024workarena,
  title={{WorkArena}: How Capable Are Web Agents at Solving Common Knowledge Work Tasks?},
  author={Drouin, Alexandre and Gasse, Maxime and Caccia, Massimo and Laradji, Issam H. and Del Verme, Manuel and Marty, Tom and Vazquez, David and Chapados, Nicolas and Lacoste, Alexandre},
  booktitle={Proceedings of the 41st International Conference on Machine Learning},
  series={Proceedings of Machine Learning Research},
  volume={235},
  pages={11642--11662},
  year={2024},
  publisher={PMLR}
}

@article{xie2024osworld,
  title={{OSWorld}: Benchmarking Multimodal Agents for Open-Ended Tasks in Real Computer Environments},
  author={Xie, Tianbao and Zhang, Danyang and Chen, Jixuan and Li, Xiaochuan and Zhao, Siheng and Cao, Ruisheng and Hua, Toh J. and Cheng, Zhoujun and Shin, Dongchan and Lei, Fangyu and others},
  journal={Advances in Neural Information Processing Systems},
  volume={37},
  pages={52040--52094},
  year={2024}
}

@inproceedings{merrill2026terminalbench,
  title={{Terminal-Bench}: Benchmarking Agents on Hard, Realistic Tasks in Command Line Interfaces},
  author={Merrill, Mike and Shaw, Alexander and Carlini, Nicholas and Li, Boxuan and Raj, Harsh and Bercovich, Ivan and Shi, Lin and Shin, Jeong and Walshe, Thomas and Buchanan, E. Kelly and others},
  booktitle={International Conference on Learning Representations},
  volume={2026},
  pages={40903--40986},
  year={2026}
}

@article{ma2024agentboard,
  title={{AgentBoard}: An Analytical Evaluation Board of Multi-turn {LLM} Agents},
  author={Ma, Chang and Zhang, Junlei and Zhu, Zhihao and Yang, Cheng and Yang, Yujiu and Jin, Yaohui and Lan, Zhenzhong and Kong, Lingpeng and He, Junxian},
  journal={Advances in Neural Information Processing Systems},
  volume={37},
  pages={74325--74362},
  year={2024}
}

@inproceedings{wu2025longmemeval,
  title={{LongMemEval}: Benchmarking Chat Assistants on Long-Term Interactive Memory},
  author={Wu, Di and Wang, Hongwei and Yu, Wenhao and Zhang, Yuwei and Chang, Kai-Wei and Yu, Dong},
  booktitle={International Conference on Learning Representations},
  volume={2025},
  pages={86809--86836},
  year={2025}
}

@article{li2026skillsbench,
  title={{SkillsBench}: Benchmarking How Well Agent Skills Work Across Diverse Tasks},
  author={Li, Xiangyi and Liu, Yimin and Chen, Wenbo and You, Bingran and Di, Zonglin and He, Yifeng and Zheng, Shenghan and Choe, Kyoung Whan and Sun, Jiankai and Wang, Shuyi and others},
  journal={arXiv preprint arXiv:2602.12670},
  year={2026}
}

@article{yang2026skillopt,
  title={{SkillOpt}: Executive Strategy for Self-Evolving Agent Skills},
  author={Yang, Yifan and Gong, Ziyang and Huang, Weiquan and Yang, Qihao and Zhou, Ziwei and Huang, Zisu and Li, Yan and Gao, Xuemei and Dai, Qi and Liu, Bei and others},
  journal={arXiv preprint arXiv:2605.23904},
  year={2026}
}

@article{ouyang2026skillos,
  title={{SkillOS}: Learning Skill Curation for Self-Evolving Agents},
  author={Ouyang, Siru and Yan, Jun and Chen, Yanfei and Han, Rujun and Wang, Zifeng and Mishra, Bhavana Dalvi and Meng, Rui and Li, Chun-Liang and Jiao, Yizhu and Zha, Kaiwen and others},
  journal={arXiv preprint arXiv:2605.06614},
  year={2026}
}

@article{li2026beyond,
  title={Beyond Semantic Similarity: Rethinking Retrieval for Agentic Search via Direct Corpus Interaction},
  author={Li, Zhuofeng and Zhang, Haoxiang and Wei, Cong and Lu, Pan and Nie, Ping and Lu, Yi and Bai, Yuyang and Feng, Shangbin and Zhu, Hangxiao and Zhong, Ming and others},
  journal={arXiv preprint arXiv:2605.05242},
  year={2026}
}

@article{huang2026rawexperience,
  title={From Raw Experience to Skill Consumption: A Systematic Study of Model-Generated Agent Skills},
  author={Huang, Zisu and Xu, Jingwen and Yang, Yifan and Gong, Ziyang and Yang, Qihao and Tian, Muzhao and Wang, Xiaohua and Lv, Changze and Gao, Xuemei and Dai, Qi and others},
  journal={arXiv preprint arXiv:2605.23899},
  year={2026}
}

@article{lin2026museautoskill,
  title={{MUSE-Autoskill}: Self-Evolving Agents via Skill Creation, Memory, Management, and Evaluation},
  author={Lin, Huawei and Li, Peng and Song, Jie and Jiang, Fuxin and Zhang, Tieying},
  journal={arXiv preprint arXiv:2605.27366},
  year={2026}
}

@article{bogavelli2025agentarch,
  title={{AgentArch}: A Comprehensive Benchmark to Evaluate Agent Architectures in Enterprise},
  author={Bogavelli, Tara and Sharma, Roshnee and Subramani, Hari},
  journal={arXiv preprint arXiv:2509.10769},
  year={2025}
}

@inproceedings{liang2025swebenchillusion,
  title={The {SWE-Bench} Illusion: When State-of-the-Art {LLMs} Remember Instead of Reason},
  author={Liang, Shanchao and Garg, Spandan and Moghaddam, Roshanak Zilouchian},
  booktitle={Proceedings of the IEEE/ACM 48th International Conference on Software Engineering: Software Engineering in Practice},
  pages={395--405},
  year={2026}
}

@article{zhang2026clawbench,
  title={{ClawBench}: Can {AI} Agents Complete Everyday Online Tasks?},
  author={Zhang, Yuxuan and Wang, Yubo and Zhu, Yipeng and Du, Penghui and Miao, Junwen and Lu, Xuan and Li, Zhuofeng and Qu, Xingwei and Guo, Zhengkang and Shen, Yuanzhe and Song, Dingjie and Zhou, Han and Zheng, Tuney and Wu, Xian and Yu, Hao and Cai, Songcheng and Lu, Yi and Hao, Yunzhuo and Lei, Minyi and Chen, Liang and Zou, Kai and Yin, Huifeng and Xu, Wendong and Jiang, Dongfu and Nie, Ping and Liu, Jiaheng and Chen, Wenhu and Allen, Kelsey R.},
  journal={arXiv preprint arXiv:2604.08523},
  year={2026}
}

@misc{openclaw2026,
  author       = {OpenClaw},
  title        = {{OpenClaw}: Your Own Personal {AI} Assistant},
  year         = {2026},
  publisher    = {GitHub},
  url          = {https://github.com/openclaw/openclaw},
  note         = {MIT License. Accessed 2026-04-21}
}

@misc{hermes2026,
  author       = {{Nous Research}},
  title        = {Hermes Agent: The Agent That Grows With You},
  year         = {2026},
  publisher    = {GitHub},
  url          = {https://github.com/NousResearch/hermes-agent},
  note         = {MIT License. Version v2026.4.16. Accessed 2026-04-21}
}

@inproceedings{xia2025agent0,
  title={{Agent0}: Unleashing Self-Evolving Agents from Zero Data via Tool-Integrated Reasoning},
  author={Xia, Peng and Zeng, Kaide and Liu, Jiaqi and Qin, Can and Wu, Fang and Zhou, Yiyang and Xiong, Caiming and Yao, Huaxiu},
  booktitle={ICLR 2026 Workshop on AI with Recursive Self-Improvement},
  year={2026},
  note={Oral},
  url={https://openreview.net/forum?id=hYYeOl58xi}
}

@article{ou2025symbolic,
  title={Symbolic learning enables self-evolving agents},
  author={Ou, Yixin and Zhou, Wangchunshu and Ding, Shengwei and Li, Long and Wu, Jialong and Wang, Tiannan and Chen, Jiamin and Wang, Shuai and Xu, Xiaohua and Zhang, Ningyu and others},
  journal={AI Open},
  volume={6},
  pages={314--322},
  year={2025},
  publisher={Elsevier},
  doi={10.1016/j.aiopen.2025.11.004}
}

@inproceedings{
khattab2024dspy,
title={{DSP}y: Compiling Declarative Language Model Calls into State-of-the-Art Pipelines},
author={Omar Khattab and Arnav Singhvi and Paridhi Maheshwari and Zhiyuan Zhang and Keshav Santhanam and Sri Vardhamanan A and Saiful Haq and Ashutosh Sharma and Thomas T. Joshi and Hanna Moazam and Heather Miller and Matei Zaharia and Christopher Potts},
booktitle={The Twelfth International Conference on Learning Representations},
year={2024},
url={https://openreview.net/forum?id=sY5N0zY5Od}
}

@article{yuksekgonul2025optimizing,
  title={Optimizing Generative {AI} by Backpropagating Language Model Feedback},
  author={Yuksekgonul, Mert and Bianchi, Federico and Boen, Joseph and Liu, Sheng and Lu, Pan and Huang, Zhi and Guestrin, Carlos and Zou, James},
  journal={Nature},
  volume={639},
  number={8055},
  pages={609--616},
  year={2025},
  publisher={Nature Publishing Group UK London}
}

@article{agarwal2024many,
  title={Many-shot in-context learning},
  author={Agarwal, Rishabh and Singh, Avi and Zhang, Lei and Bohnet, Bernd and Rosias, Luis and Chan, Stephanie and Zhang, Biao and Anand, Ankesh and Abbas, Zaheer and Nova, Azade and others},
  journal={Advances in Neural Information Processing Systems},
  volume={37},
  pages={76930--76966},
  year={2024}
}

@article{ren2026selfimprovements,
  title   = {Self-Improvements in Modern Agentic Systems: A Survey},
  author  = {Ren, Zhe and Chen, Yimeng and Guo, Dandan and Rong, Guowei and Li, Tonghui and Xiong, R. B. and Lan, Qingfeng and Wang, Wenyi and Li, Nanbo and Yang, Yibo and Zhuge, Mingchen and Schmidhuber, J{\"u}rgen},
  journal = {arXiv preprint arXiv:2607.13104},
  year    = {2026}
}

@article{lee2026recursive,
  title={Recursive Harness Self-Improvement},
  author={Lee, Hyunin and Xu, Jinglue and Seely, Jeffrey and Lee, Donghyun and Zaharia, Matei and Tang, Yujin},
  journal={arXiv preprint arXiv:2607.15524},
  year={2026}
}

@article{qu2024recursive,
  title={Recursive introspection: Teaching language model agents how to self-improve},
  author={Qu, Yuxiao and Zhang, Tianjun and Garg, Naman and Kumar, Aviral},
  journal={Advances in Neural Information Processing Systems},
  volume={37},
  pages={55249--55285},
  year={2024}
}

@inproceedings{yin2025godel,
  title={G{\"o}del agent: A self-referential agent framework for recursively self-improvement},
  author={Yin, Xunjian and Wang, Xinyi and Pan, Liangming and Lin, Li and Wan, Xiaojun and Wang, William Yang},
  booktitle={Proceedings of the 63rd Annual Meeting of the Association for Computational Linguistics (Volume 1: Long Papers)},
  pages={27890--27913},
  year={2025}
}

@article{wang2026openclaw,
  title={{OpenClaw-RL}: Train Any Agent Simply by Talking},
  author={Wang, Yinjie and Chen, Xuyang and Jin, Xiaolong and Wang, Mengdi and Yang, Ling},
  journal={arXiv preprint arXiv:2603.10165},
  year={2026}
}

@article{gao2025survey,
  title={A Survey of Self-Evolving Agents: What, When, How, and Where to Evolve on the Path to Artificial Super Intelligence},
  author={Gao, Huan-ang and Geng, Jiayi and Hua, Wenyue and Hu, Mengkang and Juan, Xinzhe and Liu, Hongzhang and Liu, Shilong and Qiu, Jiahao and Qi, Xuan and Wu, Yiran and others},
  journal={arXiv preprint arXiv:2507.21046},
  year={2025}
}

@inproceedings{sarukkai2025selfgenerated,
  title     = {Self-Generated In-Context Examples Improve {LLM} Agents for Sequential Decision-Making Tasks},
  author    = {Sarukkai, Vishnu and Xie, Zhiqiang and Fatahalian, Kayvon},
  booktitle = {Advances in Neural Information Processing Systems},
  volume    = {38},
  year      = {2025}
}

@article{zhang2026memrl,
  title   = {{MemRL}: Self-Evolving Agents via Runtime Reinforcement Learning on Episodic Memory},
  author  = {Zhang, Shengtao and Wang, Jiaqian and Zhou, Ruiwen and Liao, Junwei and Feng, Yuchen and Li, Zhuo and Zheng, Yujie and Zhang, Weinan and Wen, Ying and Li, Zhiyu and Xiong, Feiyu and Qi, Yutao and Tang, Bo and Wen, Muning},
  journal = {arXiv preprint arXiv:2601.03192},
  year    = {2026}
}

@article{fang2025comprehensive,
  title={A Comprehensive Survey of Self-Evolving {AI} Agents: A New Paradigm Bridging Foundation Models and Lifelong Agentic Systems},
  author={Fang, Jinyuan and Peng, Yanwen and Zhang, Xi and Wang, Yingxu and Yi, Xinhao and Zhang, Guibin and Xu, Yi and Wu, Bin and Liu, Siwei and Li, Zihao and others},
  journal={arXiv preprint arXiv:2508.07407},
  year={2025}
}

@inproceedings{maharana2024evaluating,
  title={Evaluating Very Long-Term Conversational Memory of {LLM} Agents},
  author={Maharana, Adyasha and Lee, Dong-Ho and Tulyakov, Sergey and Bansal, Mohit and Barbieri, Francesco and Fang, Yuwei},
  booktitle={Proceedings of the 62nd Annual Meeting of the Association for Computational Linguistics (Volume 1: Long Papers)},
  pages={13851--13870},
  year={2024}
}

@misc{berkeleyRDI2026brokenbenchmarks,
  author       = {{Berkeley RDI}},
  title        = {How We Broke Top {AI} Agent Benchmarks: And What Comes Next},
  year         = {2026},
  howpublished = {Blog post, Berkeley Center for Responsible
                  Decentralized Intelligence},
  url          = {https://rdi.berkeley.edu/blog/trustworthy-benchmarks-cont/},
  note         = {Accessed 2026-04-21}
}

@inproceedings{mialon2023gaia,
  title={{GAIA}: A Benchmark for General {AI} Assistants},
  author={Mialon, Gr{\'e}goire and Fourrier, Cl{\'e}mentine and Wolf, Thomas and LeCun, Yann and Scialom, Thomas},
  booktitle={International Conference on Learning Representations},
  volume={2024},
  pages={9025--9049},
  year={2024}
}

@inproceedings{buening2026aligning,
  title={Aligning language models from user interactions},
  author={Buening, Thomas Kleine and H{\"u}botter, Jonas and P{\'a}sztor, Barna and Shenfeld, Idan and Ramponi, Giorgia and Krause, Andreas},
  booktitle={ICML 2026 Workshop on Continual Adaptation at Scale: Towards Sustainable AI},
  year={2026},
  note={Oral},
  url={https://openreview.net/forum?id=HL3iloEczS}
}

@misc{agentzero2026,
  author       = {{Agent Zero Contributors}},
  title        = {{Agent Zero}: Personal, Organic Agentic Framework},
  year         = {2026},
  url          = {https://github.com/agent0ai/agent-zero},
  note         = {Accessed 2026-04-24}
}

@misc{zeroclaw2026,
  author       = {{ZeroClaw Labs}},
  title        = {{ZeroClaw}: Fast, Small, and Fully Autonomous {AI} Personal Assistant Infrastructure},
  year         = {2026},
  url          = {https://github.com/zeroclaw-labs/zeroclaw},
  note         = {Accessed 2026-04-24}
}

@misc{nanobot2026,
  author       = {{HKUDS}},
  title        = {{nanobot}: The Ultra-Lightweight Personal {AI} Agent},
  year         = {2026},
  url          = {https://github.com/HKUDS/nanobot},
  note         = {Accessed 2026-04-24}
}

@inproceedings{mem02026,
  title={Mem0: Building Production-Ready {AI} Agents with Scalable Long-Term Memory},
  author={Chhikara, Prateek and Khant, Dev and Aryan, Saket and Singh, Taranjeet and Yadav, Deshraj},
  booktitle={ECAI 2025},
  series={Frontiers in Artificial Intelligence and Applications},
  volume={413},
  pages={2993--3000},
  year={2025},
  publisher={IOS Press},
  doi={10.3233/FAIA251160}
}

@misc{langgraphdeepagents2026,
  author       = {{LangChain}},
  title        = {{Deep Agents} Memory Documentation},
  year         = {2026},
  url          = {https://docs.langchain.com/oss/python/deepagents/memory},
  note         = {Accessed 2026-04-24}
}

@article{xu2026amem,
  title={{A-Mem}: Agentic Memory for {LLM} Agents},
  author={Xu, Wujiang and Liang, Zujie and Mei, Kai and Gao, Hang and Tan, Juntao and Zhang, Yongfeng},
  journal={Advances in Neural Information Processing Systems},
  volume={38},
  pages={17577--17604},
  year={2025}
}

@inproceedings{he2025traject,
  title={{TRAJECT-Bench}: A Trajectory-Aware Benchmark for Evaluating Agentic Tool Use},
  author={He, Pengfei and Dai, Zhenwei and He, Bing and Liu, Hui and Tang, Xianfeng and Lu, Hanqing and Li, Juanhui and Ding, Jiayuan and Mukherjee, Subhabrata and Wang, Suhang and Xing, Yue and Tang, Jiliang and Dumoulin, Benoit},
  booktitle={International Conference on Learning Representations},
  volume={2026},
  pages={61766--61801},
  year={2026}
}

@article{li2026atbench,
  title={{ATBench}: A Diverse and Realistic Agent Trajectory Benchmark for Safety Evaluation and Diagnosis},
  author={Li, Yu and Luo, Haoyu and Xie, Yuejin and Fu, Yuqian and Yang, Zhonghao and Shao, Shuai and Ren, Qihan and Qu, Wanying and Fu, Yanwei and Yang, Yujiu and Shao, Jing and Hu, Xia and Liu, Dongrui},
  journal={arXiv preprint arXiv:2604.02022},
  year={2026}
}

@inproceedings{hu2025evaluating,
  title={Evaluating Memory in {LLM} Agents via Incremental Multi-Turn Interactions},
  author={Hu, Yuanzhe and Wang, Yu and McAuley, Julian},
  booktitle={International Conference on Learning Representations},
  year={2026}
}

@misc{letta2025benchmarking,
  author       = {{Letta}},
  title        = {Benchmarking {AI} Agent Memory: Is a Filesystem All You Need?},
  url          = {https://www.letta.com/blog/benchmarking-ai-agent-memory},
  year         = {2025},
  month        = aug,
  note         = {Accessed 2026-04-30}
}

@article{wei2025evo,
  title={{Evo-Memory}: Benchmarking {LLM} Agent Test-Time Learning with Self-Evolving Memory},
  author={Wei, Tianxin and Sachdeva, Noveen and Coleman, Benjamin and He, Zhankui and Bei, Yuanchen and Ning, Xuying and Ai, Mengting and Li, Yunzhe and He, Jingrui and Chi, Ed H and others},
  journal={arXiv preprint arXiv:2511.20857},
  year={2025}
}

@article{zheng2025lifelongagentbench,
  title={{LifelongAgentBench}: Evaluating {LLM} Agents as Lifelong Learners},
  author={Zheng, Junhao and Cai, Xidi and Li, Qiuke and Zhang, Duzhen and Li, ZhongZhi and Zhang, Yingying and Song, Le and Ma, Qianli},
  journal={arXiv preprint arXiv:2505.11942},
  year={2025}
}

@article{sumers2023cognitive,
  title={Cognitive architectures for language agents},
  author={Sumers, Theodore and Yao, Shunyu and Narasimhan, Karthik R and Griffiths, Thomas L},
  journal={Transactions on Machine Learning Research},
  year={2023}
}

@article{squire1992declarative,
  title={Declarative and nondeclarative memory: Multiple brain systems supporting learning and memory},
  author={Squire, Larry R},
  journal={Journal of cognitive neuroscience},
  volume={4},
  number={3},
  pages={232--243},
  year={1992},
  publisher={MIT Press}
}

@book{anderson2014atomic,
  title={The atomic components of thought},
  author={Anderson, John R and Lebiere, Christian J},
  year={2014},
  publisher={Psychology Press}
}

@inproceedings{hong2023metagpt,
  title={{MetaGPT}: Meta Programming for a Multi-Agent Collaborative Framework},
  author={Hong, Sirui and Zhuge, Mingchen and Chen, Jonathan and Zheng, Xiawu and Cheng, Yuheng and Wang, Jinlin and Zhang, Ceyao and Wang, Zili and Yau, Steven Ka Shing and Lin, Zijuan and others},
  booktitle={International Conference on Learning Representations},
  volume={2024},
  pages={23247--23275},
  year={2024}
}
